\pdfoutput=1
\documentclass[conference]{IEEEtran}
\IEEEoverridecommandlockouts
\usepackage{cite}
\usepackage{amsmath,amssymb,amsfonts}
\usepackage{graphicx}
\usepackage{textcomp}
\usepackage{xcolor}

\usepackage{mathrsfs}
\usepackage{booktabs} 
\usepackage{multicol}
\usepackage{enumitem}
\usepackage{multirow}
\usepackage{hhline}
\usepackage{pifont}
\usepackage{bm}
\usepackage{array}
\usepackage{algorithm}
\usepackage{algpseudocode}
\usepackage{wasysym}
\usepackage{hyperref}
\usepackage{subfigure}
\usepackage{balance}
\usepackage{comment}

\algnewcommand{\LeftComment}[1]{\State \(\triangleright\) #1}

\newtheorem{definition}{Definition}

\newcommand{\mpc}[3]{c_{\scriptscriptstyle #1#2}^{\scriptscriptstyle #3}}

\newcommand{\conch}{ConCH}
\newcommand{\conchnc}{ConCH\_nc}
\newcommand{\conchrd}{ConCH\_rd}

\long\def\comment#1{}

\def\BibTeX{{\rm B\kern-.05em{\sc i\kern-.025em b}\kern-.08em
    T\kern-.1667em\lower.7ex\hbox{E}\kern-.125emX}}

\begin{document}

\title{Leveraging Meta-path Contexts for Classification in Heterogeneous Information Networks}


\author{\IEEEauthorblockN{Xiang Li\IEEEauthorrefmark{1},
Danhao Ding\IEEEauthorrefmark{2}, Ben Kao\IEEEauthorrefmark{2},
Yizhou Sun\IEEEauthorrefmark{3} and
Nikos Mamoulis\IEEEauthorrefmark{4}}
\IEEEauthorblockA{
\IEEEauthorrefmark{1}School of Data Science and Engineering, East China Normal University, Shanghai, China\\
\IEEEauthorrefmark{2}Department of Computer Science, The University of Hong Kong, Hong Kong\\
\IEEEauthorrefmark{3}Department of Computer Science, University of California, Los Angeles, USA\\
\IEEEauthorrefmark{4}Department of Computer Science and Engineering, University of Ioannina, Greece \\
Email: \IEEEauthorrefmark{1}xiangli@dase.ecnu.edu.cn, \IEEEauthorrefmark{2}\{dhding2, kao\}@cs.hku.hk, \IEEEauthorrefmark{3}yzsun@cs.ucla.edu, \IEEEauthorrefmark{4}nikos@cs.uoi.gr}}

\maketitle

\begin{abstract}
A heterogeneous information network (HIN) has as vertices objects of different types and as edges the relations between objects, which are also of various types.
We study the problem of classifying objects in HINs.
Most existing methods perform poorly when given scarce labeled objects as training sets,
and methods that improve classification accuracy
under such scenarios
are often computationally expensive. 
To address these problems,
we propose \conch,
a graph neural network model.
\conch\ formulates the classification problem as a multi-task learning problem that
combines semi-supervised learning with self-supervised learning to learn from both labeled and unlabeled data.
\conch\ employs meta-paths, which are sequences of object types that capture semantic relationships between objects. 
\comment{
Based on meta-paths, it
considers two sources of information for an object $x$:
(1) {\it Meta-path-based neighbors} of $x$ are retrieved and ranked, and the top-$k$ neighbors are retained. 
(2) The meta-path instances of $x$ to its selected neighbors are used to derive {\it meta-path-based contexts}. 
}
\conch\ 
co-derives object embeddings and context embeddings via graph convolution. 
It also uses the attention mechanism to fuse such embeddings. 
We conduct extensive experiments to evaluate the performance of \conch\ against other 15 classification methods.
Our results show that \conch\ 
is an effective and efficient method for HIN classification.
\end{abstract}


\section{Introduction}
\label{sec:intro}
A \emph{Heterogeneous Information Network} (HIN) is one 
whose objects are of different types
and whose edges represent different types of relations between objects.
Different from a \emph{homogeneous} network, where objects (and edges) are all of one single type,
HINs are more expressive in capturing the rich semantics of objects and their relationships in real-world applications. 
In many HINs, objects are associated with descriptive labels.
For example, authors in the bibliographic network DBLP can be labeled by research areas; 
movies in Freebase, a human-curated knowledge base, are labeled by genres.
However, 
object labeling is costly and
it has been observed that only a small portion of objects are given labels in HINs such as Freebase and Yago~\cite{LiKZH16}.
Therefore,
graph-based semi-supervised learning,
which assigns labels to unlabeled objects in graphs based on a given set of labeled objects and the graph structure,
has attracted a lot of research attention.

Conventionally, graph-based semi-supervised learning methods adopt a loss function of the form
$\mathcal{L} = \mathcal{L}_{sup} + \lambda \mathcal{L}_{reg}$,
where
$\mathcal{L}_{sup}$ is a loss measured with a set of labeled objects,
and $\mathcal{L}_{reg}$ is a graph-based regularization term.
For example,
assuming that linked objects in HINs are more likely to share the same label,
some methods~\cite{ji2010graph,wan2015graph} employ the 
graph Laplacian regularization.
However, as pointed out in~\cite{KipfW17},
the edges in a graph do not necessarily imply the similarity between the connected objects. 
For example, one can construct a network of words using WordNet data with edges representing either synonym or antonym relations. 
To address the issue,
\emph{Graph Convolutional Network} (GCN)~\cite{KipfW17} has been proposed to 
classify objects in homogeneous networks without  
an explicit graph Laplacian regularization.
It is, however, challenging to apply GCN on HINs for two reasons.
First, GCN disregards the possibly different edge types and
applies the same aggregation function for encoding all heterogenous relations between
objects~\cite{xu2019relation}. Information on the edge types is therefore not used. 
Second, in HINs, more complex semantic relations between objects are often exhibited by multi-hop paths instead of single links. 
To aggregate information from important multi-hop (or {\it path-based}) neighbors,
multiple neural-net layers have to be stacked up in GCN.
However, the exponential growth in the neighborhood size as we explore path-based relations of progressively longer path lengths
greatly increases the computation cost.
Also, it has been reported that the performance of GCN degrades when there are numerous layers. 

Recently, attempts~\cite{WangJSWYCY19,zhang2019heterogeneous,hu2020heterogeneous,zhu2020hgcn,fu2020magnn} have been made to extend GCN to classify objects in HINs.
In particular,
some works~\cite{WangJSWYCY19,fu2020magnn} leverage meta-paths.
A \emph{meta-path} is a sequence of object types that captures the semantic relation between objects in HINs.
For example,
if we denote the object types \emph{Author} and \emph{Paper} in DBLP as ``A'' and ``P'', respectively,
the meta-path \emph{Author-Paper-Author} (APA)
expresses the \emph{co-authorship} relation.
Specifically, two author objects $a_1$ and $a_2$ are {\it APA-related} if a path 
$a_1$-$p$-$a_2$
exists, where
$p$
represents a paper object and an edge here represents authorship.
The use of meta-paths identifies a set of {\it significant} path-based neighbors that are semantically related
to a given object. This in turn helps limit the neighborhood size by ignoring other less significant path-based neighbors 
and thus saves computation cost. 
However,
existing GCN models in HINs could suffer from two main problems.
On the one hand, while these methods have been shown to be effective,
most studies have been performed on large training sets.
When labeled objects are scarce,
the performance of these methods degrades (as shown in Table~\ref{tab:result}).
On the other hand,
some methods improve classification accuracy by sacrificing efficiency.
For example,
HGT~\cite{hu2020heterogeneous} proposes a Transformer~\cite{VaswaniSPUJGKP17} based aggregator
to combine information from multi-type neighbors for an object.
This aggregator achitecture introduces more learnable parameters,
which increases the training difficulty and adversely affects the model efficiency.
Therefore,
a research question arises:
\emph{Given a limited set of labeled objects,
can we develop a simple model that is both effective and efficient for classification in HINs}?

In this paper,
inspired by~\cite{hu2018leveraging,xu2019relation,fu2020magnn},
we apply GCN to HINs using meta-paths
and considering the finer information provided by the paths of a meta-path.
For example, given a path
$a_1$-$p$-$a_2$, 
we can deduce a co-authorship relation between $a_1$
and $a_2$.
We further remark that
the identity of the path, which we will call a {\it context} of the co-authorship relation, can be very influential in the classification task. 
For example, the topic (label) of paper $p$ can provide important information of the research topics (labels) of the authors.
We propose to leverage meta-path-based contexts to improve classification.
We address the problems mentioned above from three perspectives. 
First,
we leverage self-supervised learning
to learn from massive unlabeled data.
The learned data-driven prior knowledge
could provide supplementary information for the shortage of labeled objects in the training set.
Specifically, 
we formulate the problem as a multi-task learning problem,
where the self-supervised loss acts as a regularizer to the supervised loss w.r.t. labeled objects.
Second, we improve modeling efficiency by 
selecting the most informative meta-path-specified neighbors for an object,
which effectively filters less-informative neighbors.
Third, we propose to compute {\it context embeddings} and integrate them into the computation of object embeddings. 
Specifically, we propose a process that performs mutual updates of 
these two kinds of embeddings.
Our main contributions are summarized as follows.

\noindent$\bullet$
We propose \conch,
which
leverages meta-path-based \textbf{Con}texts to effectively and efficiently
\textbf{C}lassify objects in \textbf{H}INs,
given scarce labeled objects as training sets.

\noindent$\bullet$
We formulate the classification problem in HINs as a multi-task learning problem that 
combines semi-supervised learning with self-supervised learning.
In particular,
when labeled objects are scarce,
self-supervision learned from unlabeled data
helps improve model performance.

\noindent$\bullet$
We improve model efficiency by putting forward a filtering strategy that selects the most informative path-based neighbors of an object.
We aggregate information from selected neighbors to compute an object's embedding.


\noindent$\bullet$
We conduct extensive experiments 
to evaluate \conch\ against $15$ other competitors. 
Our results show that 
\conch\ is highly effective and also efficient.

\comment{
 The rest of the paper is organized as follows.
 Section~\ref{sec:related} reviews related work.
 Section~\ref{sec:pre} describes related concepts and gives a formal problem definition.
 Section~\ref{sec:algorithm} presents \conch. 
 Section~\ref{sec:exp} describes our experiments and presents experimental results.
 Finally, Section~\ref{sec:conclusion} concludes the paper.\looseness=-1
}

\section{Related Work}
\label{sec:related}
We discuss four categories of related works:
graph-based semi-supervised learning, network embedding,
self-supervised learning and graph neural networks.
For a survey of heterogeneous network representation learning,
see~\cite{dong2020heterogeneous}.

\textbf{[Graph-based semi-supervised learning]}.
Semi-supervised learning has been widely studied in graphs~\cite{zhou2004learning,belkin2006manifold,zhu2003semi}. 
Generally,
a predictive score $f(l_j|x)$ of an object $x$ and a label $l_j$ is defined to indicate the likelihood that object $x$'s label is $l_j$.
The objective is to learn $f()$ for all objects and labels by minimizing a loss value that consists of two components:
(1) for any labeled object $x$,
the difference between the predictive score $f(l_j|x)$ and $x$'s true label's value, which is 1 if $x$'s label is $l_j$; 0 otherwise;
(2) for any two linked objects $x_u$ and $x_v$ in the graph,
the difference between their predictive scores $f(l_j|x_u)$ and $f(l_j|x_v)$ over the set of labels. 
The minimization of the second component is mostly achieved by explicit forms of graph Laplacian regularization.
Semi-supervised learning in HINs has also been studied ~\cite{ji2010graph,jiang2017semi,wan2015graph}.
For example,
GNetMine~\cite{ji2010graph} is a transductive classifier that applies the idea of label predictions via predictive scores $f(l_j|x)$.
As another example,
Grempt~\cite{wan2015graph} is a transductive regression model that employs meta-paths to
deduce objects' relations and it employs graph Laplacian regularization. 


\textbf{[Network embedding]}.
Network embedding has been widely studied to learn representation vectors of objects in networks~\cite{wang2016structural,ou2016asymmetric,liu2017semantic}.
These representation vectors can be fed into a variety of downstream tasks, such as link prediction, classification and recommendation.
Early studies are for homogeneous networks~\cite{PerozziAS14,GroverL16,ou2016asymmetric} with a focus of encoding a network's connectivity.
For example, 
inspired by the idea of word-context pairs in sentences from word2vec~\cite{mikolov2013distributed},
DeepWalk~\cite{PerozziAS14} introduces node-context pairs in random walk sequences. 
The random walk sequences are fed into the SkipGram model~\cite{mikolov2013distributed} to generate node embeddings. 
Other random-walk-based methods have also been proposed,
such as node2vec~\cite{GroverL16}.
Some works consider information other than network structure, 
such as object attributes~\cite{liao2018attributed} and object labels~\cite{YangCS16}.
There are also methods that study various characteristics of network embeddings.
For example,
\cite{qiu2018network}
puts forward a unified matrix factorization framework for some well-known network embedding methods,
such as LINE~\cite{tang2015line}.
Network embedding in HINs has also attracted great attention recently~\cite{shi2018easing,shi2018aspem,chang2015heterogeneous}. 
Some of these methods are meta-path-based.
For example,
metapath2vec~\cite{DongCS17} performs meta-path-based random walks to construct heterogeneous neighbors of an object.
Taking object embeddings as parameters, 
HIN2Vec~\cite{fu2017hin2vec} constructs a binary classifier that predicts 
whether a given pair of objects are related by a meta-path relation.
There are also specialized methods that are designed for specific tasks~\cite{chen2017task} or specific categories of HINs~\cite{tang2015pte}.

\textbf{[Self-supervised learning]}.
Self-supervised learning
plays an increasing role in utilizing unlabeled data.
It has been widely used in computer vision~\cite{he2020momentum,chen2020simple,wu2018unsupervised} and
natural language processing~\cite{devlin2018bert,yang2019xlnet,radford2019language}.
Recently, there are also works~\cite{you2020does,hu2020gpt,qiu2020gcc,velickovic2019deep,sun2019infograph,hu2019strategies,hassani2020contrastive,ren2019heterogeneous} 
that 
study self-supervised learning in graphs and further
apply it for network representation learning.
For example,
deep graph infomax (DGI)~\cite{velickovic2019deep}
learns node representations by contrasting node and graph encodings.
In~\cite{hassani2020contrastive},
MVGRL learns
node and graph representations
by contrasting encodings from first-order neighbors (adjacency matrix) and a graph diffusion matrix,
which are respectively taken as local and global views of a graph structure.
Further,
heterogeneous deep graph infomax (HDGI)~\cite{ren2019heterogeneous} extends DGI to HINs,
to learn node representations by mutual information maximization.

\textbf{[Graph neural networks]}. 
Kipf and Welling~\cite{KipfW17} point out that traditional semi-supervised learning approaches on graphs
rely on the assumption that edges encode similarity between objects, which may not be true.
They extend the convolution operation on graphs to encode network structures,
and propose the
GCN model to avoid explicit forms of graph Laplacian regularization.
GCN is based on spectral graph convolutional neural networks~\cite{BrunaZSL13,DefferrardBV16},
which decompose graph signals via graph Fourier transform and convolve on the spectral components.
There are also a series of spatial graph convolutional neural network models
that directly convolve information from spatially nearby neighbors of objects in graphs.
For example,
GraphSAGE~\cite{HamiltonYL17} generates an object's embedding vector by
aggregating information from a fixed-size neighborhood of the object.
Graph attention networks (GATs)~\cite{VelickovicCCRLB18}
employ the attention mechanism to learn the importance of an object's neighbors
and aggregate information from these neighbors with their learned weights.
However, all these models are designed for homogeneous networks.

There are also works that extend graph convolutional neural networks to HINs~\cite{WangJSWYCY19,hu2018leveraging,xu2019relation,zhang2019heterogeneous,hu2020heterogeneous,zhu2020hgcn,fu2020magnn,zhang2018deep,cen2019representation,yun2019graph,yang2020domain}. 
For example,
to generate an object's embedding,
HetGNN~\cite{zhang2019heterogeneous} first derives a set of object-type-based neighbor embeddings by
aggregating information from neighbors in the same type
with bi-directional LSTM.
After that,
HetGNN adopts the vanilla attention mechanism to fuse 
these object-type-based neighbor embeddings.
Inspired by the architecture of Transformer~\cite{VaswaniSPUJGKP17},
HGT~\cite{hu2020heterogeneous} proposes heterogeneous graph transformer that leverages multi-head attentions
to distinguish different relation types between an object and its neighbors.
Further,
HGCN~\cite{zhu2020hgcn} distinguishes the various edge types and derives multiple sub-networks.
In each convolutional layer,
for each object type in each sub-network,
HGCN aggregates information from an object's neighbors 
by leveraging multiple kernels that use different convolutional filters with different aggregation strategies.
Embeddings derived from various kernels in these sub-networks are fused to construct the object's relational feature vector.
Finally, relational features and original features of the object are concatenated,
which are fed into a MLP to predict the object's labels.
However, with more neural-net layers stacked up in these models,
the neighborhood size of an object exponentially increases.
To address the problem,
some methods use meta-paths to identify a subset of {\it multi-hop neighbors} of a given object\footnote{
A multi-hop neighbor of an object $x$ is one that is reached from $x$ via a path of multiple edges.}.
For example,
for each given meta-path $\mathcal{P}$ and an object $x$,
HAN~\cite{WangJSWYCY19} uses a node-level attention to learn the importance of all multi-hop neighbors specified by $\mathcal{P}$
and generates $x$'s embedding by aggregating information from these neighbors.
After that, a meta-path-level attention is used to learn the importance of meta-paths, and $x$'s embeddings based on the various meta-paths
are fused to obtain $x$'s final embedding.
MAGNN~\cite{fu2020magnn} further improves HAN by utilizing meta-path-based contexts.
Given a meta-path $\mathcal{P}$ and an object $x$,
MAGNN learns the importance of path instances of $\mathcal{P}$ starting from $x$ and 
generates $x$'s embedding by 
aggregating information from these path instances with the learned weights.
Finally, the embeddings learned from various meta-paths are combined to generate $x$'s final embedding.


Despite their success,
we observe that 
most existing methods perform poorly when the number of given labeled objects is limited.
Also,
many of them improve the model's effectiveness by sacrificing efficiency.
Different from these methods,
our proposed method \conch\ 
can achieve superior performance with high efficiency, given scarce labeled objects.



\section{Definitions}
\label{sec:pre}

In this section we formally define various concepts and the HIN classification problem.

\begin{definition}
\label{def:hin}
\textbf{Heterogeneous Information Network (HIN)}. 
Let $\mathcal{T}$ = $\{T_1, ..., T_m \}$ be a set of $m$ object types.
For each type $T_i$, 
let $\mathcal{X}_i$ be the set of objects of type $T_i$.
An HIN is a graph $\mathcal{G} = (\mathcal{V}, \mathcal{E})$, where
$\mathcal{V} = \bigcup_{i=1}^m\mathcal{X}_i$ is a set of nodes and
$\mathcal{E}$ is a set of edges, each represents a binary relation between two objects in $\mathcal{V}$.
An HIN $\mathcal{G} = (\mathcal{V}, \mathcal{E})$ is associated with two mappings. 
(1) $\phi: \mathcal{V} \rightarrow \mathcal{T}$ is an object-type mapping that maps an object in $\mathcal{V}$ into its type, and
(2) $\psi: \mathcal{E} \rightarrow \mathcal{R}$ is a link-relation mapping that maps a link in $\mathcal{E}$ into a relation in a set of 
relations $\mathcal{R}$.
A network is an HIN if $|\mathcal{T}| > 1$ or $|\mathcal{R}| > 1$; otherwise, the network is homogeneous.
\hfill$\Box$
\end{definition}

\begin{definition}
\textbf{Network schema}.
The network schema of an HIN $\mathcal{G}$, denoted by $T_\mathcal{G} = (\mathcal{T}, \mathcal{R})$,
shows how objects of different types are related by the relations in $\mathcal{R}$. 
A {\it schematic graph} is used to represent $T_\mathcal{G}$ with $\mathcal{T}$ and $\mathcal{R}$ being the node set and the edge set, 
respectively.
Specifically, there is an edge ($T_i$, $T_j$) in the schematic graph iff there is a relation in $\mathcal{R}$ that relates
objects of type $T_i$ to objects of type $T_j$.
\hfill$\Box$
\end{definition}

\begin{figure}[!htbp]
    \centering
        \includegraphics[width = 0.6\linewidth]{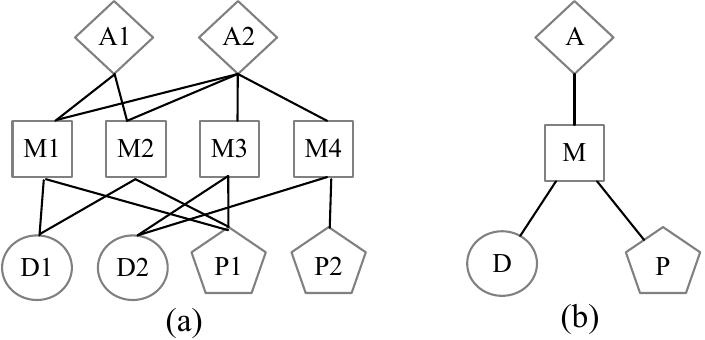}
        \caption{An HIN and its schematic graph}
        \label{figure:example}
\end{figure}

As an example,
Fig.~\ref{figure:example}(a) shows a movie HIN with object types: $\mathcal{T}$ = 
\{Movie($\Box$), Actor($\Diamond$), Director($\Circle$), Producer($\pentagon$)\}.
The set $\mathcal{R}$ consists of three relations, which are illustrated by the three edges in the schematic
graph (Fig.~\ref{figure:example}(b)). 
For example,
the edge A\textendash M in Fig.~\ref{figure:example}(b) represents the relation that an (A)ctor acts in a (M)ovie,
and the edge M\textendash D denotes the relation that a (M)ovie is directed by a (D)irector.

\begin{definition}
\textbf{Meta-path}.
A meta-path $\mathcal{P}$: $T_1\stackrel{R_1}{\longrightarrow} \cdots\stackrel{R_l}{\longrightarrow} T_{l+1}$
 defines a composite relation $R = R_1\circ \cdots \circ R_l$ 
 that relates objects of type $T_1$ to objects of type $T_{l+1}$.
 If two objects $x_u$ and $x_v$ are related by the composite relation $R$,
 then there is a path, denoted by $p_{x_u\leadsto{x_v}}$, that connects $x_u$ to $x_v$ in $\mathcal{G}$.
 Moreover, the sequence of objects and links in $p_{x_u\leadsto{x_v}}$ matches the sequence of types $T_1$, ..., $T_{l+1}$ and 
 relations $R_1$, ..., $R_l$ based on the object-type mapping $\phi$ and the link-relation mapping $\psi$, respectively.
 We say that $p_{x_u\leadsto{x_v}}$ is a {\textbf{path instance}} of $\mathcal{P}$, denoted by
 $p_{x_u\leadsto{x_v}} \vdash \mathcal{P}$.
 \hfill$\Box$
\end{definition}


In Fig.~\ref{figure:example}(a), the path $p_{M1 \leadsto M2}$ = M1 $\rightarrow$ A1 $\rightarrow$ M2
is an instance of the meta-path \emph{Movie-Actor-Movie} (abbrev. MAM),
which captures the relation between two movies that feature the same actor;
the path $p_{M2 \leadsto M3}$ = M2 $\rightarrow$ P1 $\rightarrow$ M3
is an instance of  the meta-path \emph{Movie-Producer-Movie} (abbrev. MPM),
which expresses the relation between two movies that are produced by the same producer.
Path instances describe the details on how two objects are related by a meta-path,
which can be used to define meta-path-based contexts.

\begin{definition}
\label{def:context}
\textbf{Meta-path-based context}~\cite{hu2018leveraging}.
Given two objects $x_u$ and $x_v$ that are related by a meta-path $\mathcal{P}$,
the meta-path-based context $c_{uv}^{\mathcal{P}}$ 
is the set of path instances
of $\mathcal{P}$ connecting objects $x_u$ and $x_v$,
i.e., $c_{uv}^{\mathcal{P}} = \{p_{x_u\leadsto x_v} |  \;p_{x_u\leadsto x_v} \vdash \mathcal{P}\}$.
\hfill$\Box$
\end{definition}

In the following discussion, we will use {\it mp-context} or simply {\it context} as short forms 
of ``meta-path-based context''.
As an example, in Fig.~\ref{figure:example}(a), the mp-context of M1 and M2 w.r.t. the meta-path 
MAM is
$c_{\scriptscriptstyle M_1 M_2}^{\scriptscriptstyle MAM}$ = \{M1 $\rightarrow$ A1 $\rightarrow$ M2, M1 $\rightarrow$ A2 $\rightarrow$ M2\}.

\begin{definition}
\label{def:classification}
\textbf{Classification in HINs}. Given an HIN $\mathcal{G} = (\mathcal{V}, \mathcal{E})$,
a target object type $T$ to be classified,
a label set $L= \{l_1,l_2,...,l_r\}$,
and a set of meta-paths $\mathcal{PS}$,
let $\mathcal{X} = \mathcal{Y} \cup \mathcal{U}$, where $\mathcal{Y}$ is a set of labeled objects and $\mathcal{U}$ is a set of unlabeled ones,
the classification problem in HINs is to
predict the labels of objects in $\mathcal{U}$.
 \hfill$\Box$
\end{definition}


\section{Algorithm}
\label{sec:algorithm}

\begin{figure*}[!htbp]
    \centering
        \includegraphics[width = 0.9\linewidth]{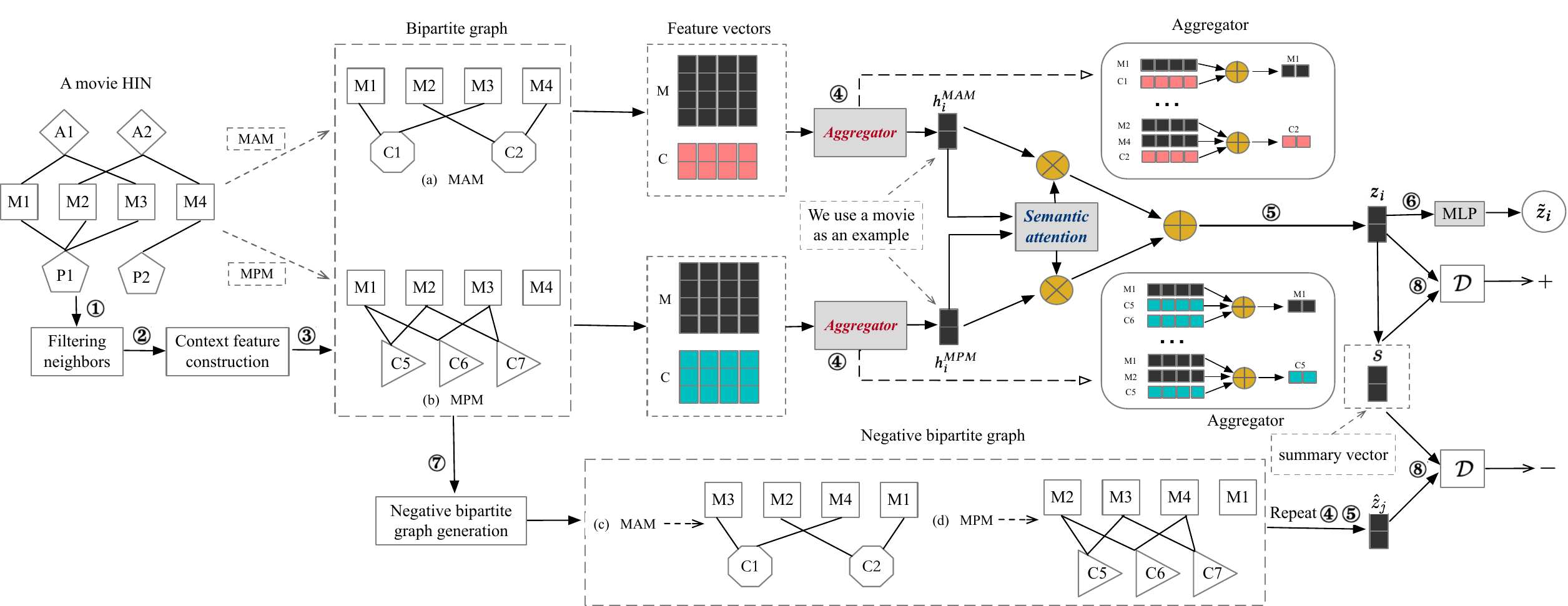}
        \caption{The overall framework of the \conch\ model. For details of each step, see Sec.~\ref{sec:algorithm}.}
        \label{figure:framework}
\end{figure*}

In this section we describe our \conch\  algorithm.
We first give an overview of \conch, which is illustrated in 
Fig.~\ref{figure:framework}. 
For an object $x$ in an HIN, \conch\ first identifies a selected set of $x$'s path-based neighbors based on
a given set of meta-paths that express various semantic relations (Step \ding{172}).
These meta-path-based relations relate objects in the HIN with path instances, which are taken as the mp-contexts (Step \ding{173}).
\conch\ then constructs a bipartite graph for each given meta-path $\mathcal{P}$ (Step \ding{174}). The bipartite graph represents objects and the mp-contexts
that relate them, and it facilitates the updating of objects' and contexts' embeddings (Step \ding{175}). 
After that, the objects' embeddings generated by various meta-paths are fused to obtain the objects' final embeddings (Step \ding{176}),
which are then fed into a two-layer MLP for label prediction (Step \ding{177}).
To better utilize unlabeled data,
\conch\ also uses self-supervised learning.
For each meta-path,
in addition to the bipartite graph constructed in Step \ding{174},
\conch\ further generates a ``negative'' bipartite graph (Step \ding{178}).
From these two graphs, 
a positive sample set and a negative sample set are constructed.
After that,
\conch\ 
uses a discriminator $\mathcal{D}$ to distinguish the two sets (Step \ding{179}).
Finally,
our objective function is formulated as a multi-task learning problem by combing both the supervised loss and the self-supervised loss.
Next, we describe each component in detail.

\subsection{Neighbor filtering}
\label{subsec:fn}
Given an object $x$ in an HIN, our objective is to identify other objects in the network that are similar to $x$ and use their labels
to infer that of $x$. We call these other objects {\it relevant neighbors}. 
Note that these neighbors could be multiple hops away from $x$, especially if they are related to $x$ via paths that are instances of
certain given meta-paths that express important semantic relations between objects. 
An interesting question is how relevant neighbors can be obtained effectively and efficiently. 
If meta-paths are numerous (e.g., those that are obtained via automatic methods) and long, then the neighbors derived from
them could cover a large scope of the network.  
Directly aggregating information from large numbers of neighbors would make model construction inefficient.
While there are sampling methods~\cite{HamiltonYL17} of neighbors, the sampling process itself could be time-consuming and less
relevant neighbors may be sampled instead of the more relevant ones. Our first step is to filter neighbors and select the most relevant
neighbors of an object.


Given a meta-path $\mathcal{P}$,
\conch\
measures the similarity between two objects $x_u$ and $x_v$ of the same type w.r.t. $\mathcal{P}$ by \emph{PathSim}~\cite{SunHYYW11}:
\begin{equation}
\small
\label{eq:pathsim}
\text{PS}(x_u,x_v) = \frac{2\times\vert\{ p_{x_u \leadsto x_v}|p_{x_u \leadsto x_v} \vdash \mathcal{P} \}\vert}{\vert\{ p_{x_u \leadsto x_u}|p_{x_u \leadsto x_u} \vdash \mathcal{P} \}\vert +\vert\{ p_{x_v \leadsto x_v}|p_{x_v \leadsto x_v} \vdash \mathcal{P} \}\vert }.
\end{equation}
PathSim has been shown to be very effective in a variety of data mining tasks in HINs~\cite{li2017semi,wan2015graph,luo2014hetpathmine,LiKZH16}.
Given an object $x$ and a meta-path $\mathcal{P}$, \conch\ collects a set of neighbors $N_x^{\mathcal{P}}$ that contains the 
top-$k$ objects with the highest PathSim scores with $x$ w.r.t. $\mathcal{P}$.
Using only the top-$k$ neighbors significantly reduces the number of relevant objects that need to be considered in the subsequent steps of the
classification process. Also, it removes less relevant ones and thus helps reduce noise.
This further improves the classification effectiveness and efficiency. 


\subsection{Context feature construction}
\label{sec:contextcon}
Recall that given two objects $x_1$ and $x_l$, and a meta-path $\mathcal{P}$, the mp-context $\mpc{1}{l}{\mathcal{P}}$
is a set of path instances of $\mathcal{P}$ between $x_1$ and $x_l$. 
Learning a context embedding from scratch by
treating it as learnable parameter is very costly. 
To reduce the number of parameters,
we construct a feature vector for each context.
Specifically, we apply a conventional HIN embedding method,
such as \emph{metapath2vec}~\cite{DongCS17}, to obtain initial object embeddings.
Then, we generate a path instance's embedding by aggregating the embedding vectors of all the objects in the path instance. 
Given 
a path instance $p_{x_1\leadsto{x_l}} = x_1x_2\ldots x_{l} \vdash \mathcal{P}$, 
its embedding is computed by
\begin{equation}
o_{p_{x_1\leadsto{x_{l}}}} = \text{MEAN}(\{e_i\}_{i=1}^{l}),
\end{equation}
where $e_i$ is the initial embedding vector of $x_i, \forall 1 \leq i \leq l$.
An {\it initial} context embedding is further obtained by aggregating path instances' embeddings:
\begin{equation}
\label{eq:context_feature}
o_{c_{1l}^{\mathcal{P}}} = \text{MEAN}(\{o_{p_{x_1\leadsto{x_{l}}}}\} \vert p_{x_1\leadsto{x_{l}}} \in c_{1l}^{\mathcal{P}}),
\end{equation}
which serves as the context $\mpc{1}{l}{\mathcal{P}}$'s 
feature vector.

\subsection{Meta-path-based bipartite graphs}
\label{subsec:pg}
In the previous steps,
we obtained a selected set of path-based neighbors of objects and mp-contexts.
We next update their embeddings through a mutual update process.
Note that a meta-path-based relation between two objects is influenced by the meta-path instances (i.e., a context) that connect the two objects;
At the same time, a context is influenced by the objects that compose the meta-path instances. Hence, we update object embeddings and context
embeddings together. 
To facilitate such an update, we first construct, for each meta-path $\mathcal{P}$, a bipartite graph $G_{\mathcal{P}} = (\mathcal{X}, V_C^{\mathcal{P}}, E_{OC}^{\mathcal{P}})$,
where $\mathcal{X}$ is the set of objects to be classified,
$V_C^{\mathcal{P}}$ is the set of meta-path contexts derived from $\mathcal{P}$,
and $E_{OC}^{\mathcal{P}}$ is the set of edges such that $e_{ij}$ connects object $x_i$ in $\mathcal{X}$ and context $c_j$ in $V_C^{\mathcal{P}}$
if the path instances in $c_j$ start from or end at $x_i$.
Note that our neighbor filtering scheme (see Section~\ref{subsec:fn}) derives top-$k$ neighbors for each object. Therefore, the degree
of any object $x$ in the bipartite graph is at most $k$. 


Fig.~\ref{figure:framework} shows example bipartite graphs derived from a movie HIN.
For the meta-path Movie-Actor-Movie (MAM),
M1 and M3 are connected by the path instance M1$\rightarrow$ A1 $\rightarrow$ M3, and 
M2 and M4 are connected by M2$\rightarrow$ A2 $\rightarrow$ M4.
The bipartite graph for MAM is shown in Fig.~\ref{figure:framework}(a),
in which context C1 = \{M1$\rightarrow$ A1 $\rightarrow$ M3\} and C2 = \{M2$\rightarrow$ A2 $\rightarrow$ M4\}.


We update object embeddings (denoted by $h_x^{\mathcal{P}}$) and context embeddings (denoted by $h_c^{\mathcal{P}}$) from each other. 
For each context $c_j$ that is linked to two objects $x_u$ and $x_v$ in $G_\mathcal{P}$,
we update its embedding vector $h_{c_j}^{\mathcal{P}}$ in the $(t+1)$-st timestep by aggregating information from both $x_u$ and $x_v$:
\begin{equation}
\small
\label{eq:upd_node}
h^{\mathcal{P},(t+1)}_{c_j} = \text{ReLU}\left( W_1^{(t)}\cdot h^{\mathcal{P},(t)}_{x_u} + W_1^{(t)} \cdot h^{\mathcal{P},(t)}_{x_v} + W_2^{(t)}\cdot h^{\mathcal{P},(t)}_{c_j}\right),
\end{equation} 
where $W_1^{(t)}$  and $W_2^{(t)}$ are two linear transformation matrices to be learned.
For any object $x_i \in \mathcal{X}$, 
it connects to at most $k$ contexts.
From these contexts,
we update the embedding vector of $x_i$ in the $(t+1)$-st timestep by:
\begin{equation}
\label{eq:upd_context}
h^{\mathcal{P},(t+1)}_{x_i} = \text{ReLU}( W_3^{(t)} \cdot \sum_{j} h^{\mathcal{P},(t)}_{c_j} + W_4^{(t)}\cdot h^{\mathcal{P},(t)}_{x_i}),
\end{equation} 
where $W_3^{(t)}$ and $W_4^{(t)}$ are another two weight matrices.
Here, we use the \emph{sum} aggregator.
It is efficient and since we have retained only the top-$k$ neighbors that are the most informative, the aggregation should also be effective.

\subsection{Semantic aggregation}
\label{sc:sa}
Our next step is to fuse together the object embeddings derived from different meta-paths.
Since different meta-paths represent different semantic relations, they 
may contribute in different extents in the classification task. 
We use the vanilla attention mechanism to learn the importance of meta-paths.
Given a meta-path $\mathcal{P}$ and the embedding vector $h_{x_i}^{\mathcal{P}}$ of an object $x_i$,
we feed $h_{x_i}^{\mathcal{P}}$ into a two-layer MLP and get:\looseness=-1
\begin{equation}
\label{eq:mp_mlp}
\tilde{w}_{x_i}^{\mathcal{P}} = a^T \cdot \xi (W_5 \cdot \text{tanh}(W_6 \cdot h^{\mathcal{P}}_{x_i})),
\end{equation}
where $a$ is the weight vector, $\xi$ is the leaky\_ReLU function, and $W_5$ and $W_6$ are two linear transformation matrices.
We regularize $\tilde{w}_{x_i}^{\mathcal{P}}$ by softmax function 
and compute the attention weight $w_{x_i}^{\mathcal{P}}$ of the meta-path $\mathcal{P}$ by
\begin{equation}
\label{eq:mp_soft}
w_{x_i}^{\mathcal{P}} = \text{softmax}(\tilde{w}_{x_i}^{\mathcal{P}}) = \dfrac{\text{exp}(\tilde{w}_{x_i}^{\mathcal{P}})}{\sum_q \text{exp}(\tilde{w}_{x_i}^{\mathcal{P}_q})}.
\end{equation}
Finally,
we aggregate the embedding vectors across meta-paths
and generate the final embedding vector $z_i$ of $x_i$ by:
\begin{equation}
\label{eq:mp_asum}
z_i = \text{ReLU}(\sum_q w_{x_i}^{\mathcal{P}_q} \cdot h^{\mathcal{P}_q}_{x_i}).
\end{equation}

\subsection{The \conch\ model}
After the semantic aggregation, 
a standard way to guide parameter learning is to use label information in the training set.
Specifically,
we first feed $z_i$ in Eq.~\ref{eq:mp_asum} into a two-layer MLP
to output a label vector $\tilde{z}_i$ of length $r$ ($r$ is the number of labels):
\begin{equation}
\label{eq:loss_mlp}
\tilde{z}_i= W_7(\text{ReLU}(W_8 \cdot z_i)),
\end{equation}
where $W_7$ and $W_8$ are learnable weight matrices.
The label corresponding to the largest-value entry in $\tilde{z}_i$ is taken as the predicted label of $x_i$.
Then we minimize the Cross-Entropy loss function:
\begin{equation}
\label{eq:loss}
\mathcal{L}_{sup} =   - \sum_{x_i \in \mathcal{Y}} y_i \ln(\tilde{z}_i), 
\end{equation}
where $\mathcal{Y}$ is the training set of labeled objects and
$y_i$ is the one-hot encoded true label vector of object $x_i$.
While the supervised loss in Eq.~\ref{eq:loss} is useful,
the model performance could be adversely affected when the number of labeled objects is limited.
Further,
self-supervised learning
has shown great potential in utilizing unlabeled data.
Therefore, 
we incorporate self-supervised learning into the learning process 
to leverage both labeled and unlabeled objects. 

After the final embedding vector of an object is derived (in Eq.~\ref{eq:mp_asum}),
we compute a summary vector $s$ as the global representation encoder,
which retains the information from all the objects of the type to be classified.
We simply take the averaging operator and get: 
\begin{equation}
\label{eq:global_vector}
s = \text{MEAN}(\{z_i\}_{i=1}^n),
\end{equation}
where $n$ is the number of objects to be classified.
Then we maximize the mutual information between node embeddings and the global representation,
which encourages each single object to have access to the information of others.
Following~\cite{velickovic2019deep},
we use a noise-contrastive type objective function with a standard binary cross entropy loss:
\begin{equation}
\label{eq:ss}
\scriptsize
\mathcal{L}_{ss} = \frac{1}{N+M} \left[ \sum_{i=1}^N \mathbb{E}_{Pos} [\log \mathcal{D}(z_i, s)] + \sum_{j=1}^M \mathbb{E}_{Neg} [\log (1 - \mathcal{D}(\hat{z}_j, s))] \right],
\end{equation}
where $\mathcal{D}$ is a discriminator to distinguish the positive sample set $Pos = \{(z_i,s)\}_{i=1}^N$ from the negative sample set $Neg = \{(\hat{z}_j,s)\}_{j=1}^M$.
The sample $(z_i, s)$ is positive when object $x_i$ belongs to the original graph;
the sample $(\hat{z}_j, s)$ is negative 
when object $x_j$ is the generated fake one.
$\mathcal{D}(z_i, s)$ represents the probability score assigned to the input node-summary pair, and it is defined as
\begin{equation}
\label{eq:d}
\mathcal{D}(z_i, s) = \sigma(z_i^TW_{\mathcal{D}}s),
\end{equation}
where $\sigma$ is the sigmoid function and $W_{\mathcal{D}}$ is the learnable weight matrix.
To generate negative samples,
inspired by~\cite{ren2019heterogeneous},
for each meta-path-based bipartite graph,
we fix the adjacency matrix unchanged
and randomly shuffle rows of the initial object feature matrix,
which generates a ``negative'' bipartite graph.
For example,
Fig.~\ref{figure:framework}(a) and Fig.~\ref{figure:framework}(c) show the original bipartite graph (we call it ``positive'' for comparison)
and the ``negative'' one derived by the meta-path MAM, respectively.
Feature vectors of movies in the negative bipartite graph are shuffled from that in the positive graph.
For the constructed negative bipartite graphs,
we repeat steps \ding{175} and \ding{176} (in Sec~\ref{subsec:pg} and~\ref{sc:sa})
to get $\hat{z}_j$ for each object $x_j$ and further construct $Neg$.

Finally,
to learn from both labeled and unlabeled data,
we formulate the problem as a multi-task learning problem by combining $\mathcal{L}_{sup}$ and $\mathcal{L}_{ss}$ to get a loss function:
\begin{equation}
\label{eq:obj}
\mathcal{L} = \mathcal{L}_{sup} + \lambda \mathcal{L}_{ss},
\end{equation}
where $\lambda$ controls the relative importance of the two terms.
Here,
the self-supervised loss acts as a regularizer 
that learns from unlabeled data.
The learned data-driven prior knowledge 
could provide supplementary information to the training set
when the number of labeled objects is limited.
This further enhances the model's capability.
The loss function can be optimized by stochastic gradient descent.
To prevent overfitting, 
we further regularize all the weight matrices $W_j$'s mentioned in Eqs.~\ref{eq:upd_node} -~\ref{eq:d}.


The major computation cost of \conch\ comes from the updates of object and context embeddings.
Let $n$ be the number of objects to be classified. 
In each bipartite graph,
there are at most $kn$ contexts.
Since the degree of each object is at most $k$
and that of a context is at most $2$,
the total time complexity of embedding updates is 
$O(3 k n d_1 d_2)$,
where 
$d_1$ and $d_2$ are respectively the maximum numbers of rows and columns in the transformation matrices of the MLPs.
Further,
for each meta-path, 
we generate a ``positive'' bipartite graph and a ``negative'' one.
Therefore, 
the total time complexity of embedding updates is $O(6 k n d_1 d_2 |\mathcal{PS}|)$,
where $ |\mathcal{PS}|$ is the number of meta-paths.
Since the computation for each bipartite graph is independent from the others',
\conch\ can be easily parallelized.
Finally, we summarize \conch\ in Alg.~\ref{alg}.

\begin{algorithm}
\begin{small}
\caption{\conch}
\label{alg}
\begin{algorithmic}[1]
\Require An HIN $\mathcal{G} = (\mathcal{V}, \mathcal{E})$;
target object type $T$ and object set $\mathcal{X}_T$;
a pre-defined meta-path set $\mathcal{PS}$;
object feature matrix $X$;
the number of layers $L$;
\Ensure Final object embeddings $\{z_i\}_{i=1}^{|\mathcal{X}_{T}|}$
\For {$\mathcal{P}$ in $\mathcal{PS}$}
\State Calculate $\mathcal{P}$-induced \emph{PathSim} scores based on Eq.~\ref{eq:pathsim}
\State Filter $\mathcal{P}$-based neighbors for $x_i$, $\forall x_i \in \mathcal{X}_T$
\State Construct initial context features $o_{c^{\mathcal{P}}}$ based on Eq.~\ref{eq:context_feature}
\State Construct the bipartite graph $G_{\mathcal{P}}$ 
\EndFor
\LeftComment{Lines 8-22 correspond to one epoch}
\For {$\mathcal{P}$ in $\mathcal{PS}$}
\State $\hat{X} = \text{shuffle}(X)$
\LeftComment{Let $f_i$ and $\hat{f}_i$ denote the feature vectors of object $x_i$ in $X$ and $\hat{X}$, respectively}
\State $h^{\mathcal{P},(0)}_{x_{i}} \leftarrow f_i$, $\hat{h}^{\mathcal{P},(0)}_{x_{i}} \leftarrow \hat{f}_i$, $\forall x_i \in \mathcal{X}_T$
\State $h^{\mathcal{P},(0)}_{c_{j}} \leftarrow o_{c_{j}^{\mathcal{P}}}$, $\hat{h}^{\mathcal{P},(0)}_{c_{j}} \leftarrow o_{c_{j}^{\mathcal{P}}}$, $\forall c_j \in V_C^{\mathcal{P}} $
\For {$t \leftarrow 0$ to $L-1$}
\State Update $h^{\mathcal{P},(t+1)}_{c_j}$ and $\hat{h}^{\mathcal{P},(t+1)}_{c_j}$ by Eq.~\ref{eq:upd_node}, $\forall c_j \in V_C^{\mathcal{P}} $
\State Update $h^{\mathcal{P},(t+1)}_{x_i}$ and $\hat{h}^{\mathcal{P},(t+1)}_{x_i}$ by Eq.~\ref{eq:upd_context}, $\forall x_i \in \mathcal{X}_T$
\EndFor
\EndFor
\State Compute $z_i$ and $\hat{z}_i$ by Eqs.~\ref{eq:mp_mlp} - \ref{eq:mp_asum}, $\forall x_i \in \mathcal{X}_T$
\State Construct the global vector $s$ based on Eq.~\ref{eq:global_vector} 
\State Construct the positive sample set $Pos = \{(z_i,s)\}_{i=1}^N$ and the negative sample set $Neg = \{(\hat{z}_j,s)\}_{j=1}^M$
\State Construct $\mathcal{L}_{sup}$, $\mathcal{L}_{ss}$ and $\mathcal{L}$
\State Optimize $\mathcal{L}$ to update weight matrices $W_j$'s 
\State \Return $\{z_i\}_{i=1}^{|\mathcal{X}_T|}$ 
\end{algorithmic}
\end{small}
\end{algorithm}

\section{Experimental Evaluation}
\label{sec:exp}
We evaluate \conch's effectiveness and efficiency.
We 
compare \conch\ with 15 other methods by their \emph{Micro-F1} and \emph{Macro-F1} scores.
We also show results on meta-path weight learning and parameter analysis.

\subsection{Datasets}
We use three datasets: DBLP,
Yelp,
and Freebase.
%
DBLP\footnote{https://dblp.uni-trier.de/} is a bibliographic network of academic publications.
Yelp\footnote{https://www.yelp.com/academic{\_}dataset/} contains information of businesses, such as user data and reviews.
Freebase\footnote{https://www.freebase.com/} is a knowledge base of facts recorded with the RDF model.
The three datasets are representative HINs.
We define a classification task for each dataset as follows:

\noindent{\small$\bullet$}
\textbf{DBLP}: We extracted a subset from the \emph{dblp-4area} dataset~\cite{SunYH09},
which contains 4,057 authors (A), 14,376 papers (P) and 20 conferences (C). 
Links include A-P (an author publishes a paper) and P-C (a paper is published at a conference).
Authors are classified into one of four research areas: database (DB), data mining (DM), machine learning (ML), and  information retrieval (IR). 
Following~\cite{yang2020relation},
we compute, for each author, a 300-dimension attribute vector
by averaging the word embeddings\footnote{http://nlp.stanford.edu/data/glove.840B.300d.zip}
of keyword terms of his/her published papers.
We consider the meta-path set \{APA, APAPA, APCPA\}.
Our task is to classify authors into their research areas.
The ground truth labels of authors are given in dblp-4area data,
which labels each author by his/her primary research area.

\noindent{\small$\bullet$}
\textbf{Yelp-Restaurant}: The dataset is about restaurant businesses in Yelp.
It contains 2,614 businesses (B), 33,360 reviews (R), 1,286 users (U) and 82 food relevant keywords (K).
Links include B-R (a business receives a review), U-R (a user writes a review) and K-R (a keyword is included in a review).
We classify restaurants by the types of food they provide: \emph{Fast Food}, \emph{Sushi Bars}, or \emph{American New Food}.
Each restaurant is associated with two categorical attributes: reservation (whether reservation is required) and service (whether waiter service is provided).
We use the meta-paths \{BRURB, BRKRB\}. 
The ground truth labels of restaurants are given in Yelp.

\noindent{\small$\bullet$}
\textbf{Freebase-Movie}: 
We constructed a dataset on movies from Freebase that contains 3,492 movies (M), 33,401 actors (A), 2,502 directors (D) and 4,459 producers (P).
Links include M-A (movie and its actor), M-D (movie and its director) and M-P (movie and its producer).
Our task is to classify movies by genre,
which includes \emph{Action}, \emph{Comedy} and \emph{Drama}. 
We encode a one-hot feature vector for each movie.
In our experiments,
we consider meta-paths \{MAM, MDM, MPM\}. 
We take the given genres of movies as the ground truth.

\comment{
\subsection{Measures}
\label{sec:measures}
We use three popular measures, namely, \emph{purity}, \emph{adjusted mutual information (AMI)}, and \emph{rand index (RI)}, to evaluate clustering quality~\cite{vinh2010information,lin2010power}.

Consider a clustering $\mathcal{C} = \{C_1, \ldots, C_k\}$ produced by a clustering algorithm
and a gold standard (true) clustering
$\mathcal{C}_t = \{ \hat{C}_1, \ldots, \hat{C}_k\}$.
For each cluster $C_i \in \mathcal{C}$, we find the cluster $\hat{C}_j \in \mathcal{C}_t$ that overlaps
with $C_i$ the most. 
The purity of cluster $C_i$ is the fraction of objects in $C_i$ that fall in the overlap, i.e., 
($\max_j |C_i \cap \hat{C}_j|) / |C_i|$. 
The purity of a clustering $\mathcal{C}$ is the average of its clusters' purities, weighted by the cluster sizes:
\begin{equation}
purity(\mathcal{C}_t,\mathcal{C}) = \frac{1}{n}\sum_{i}\max_{j}|C_i \cap \hat C_j|.
\end{equation}
The adjusted mutual information ({\it AMI}) is mutual information with the agreement due to chance between clusterings corrected, and
is given by,
\begin{equation}
\mathit{AMI}(\mathcal{C}_t,\mathcal{C}) = \frac{MI(\mathcal{C}_t,\mathcal{C}) - E\{MI(\mathcal{C}_t,\mathcal{C})\}}{\max\{H(\mathcal{C}_t),H(\mathcal{C})\} - E\{MI(\mathcal{C}_t,\mathcal{C})\}},
\end{equation}
where $MI(\mathcal{C}_t,\mathcal{C})$ is the mutual information between $\mathcal{C}_t$ and $\mathcal{C}$,
$H(\mathcal{C}_t)$ and $H(\mathcal{C})$ are the entropies of $\mathcal{C}_t$ and $\mathcal{C}$, respectively,
and $E\{MI(\mathcal{C}_t,\mathcal{C})\}$ is the expected mutual information between the two clusterings
$\mathcal{C}_t$ and $\mathcal{C}$.

Rand index ({\it RI}) considers object pairs in measuring clustering quality. It is defined as:
\begin{equation}
RI(\mathcal{C}_t,\mathcal{C}) = (N_{00} + N_{11}) / {\tbinom n2},
\end{equation}
where $N_{00}$ is the number of object pairs that are put into the same cluster in $\mathcal{C}_t$
as well as in the same cluster in $\mathcal{C}$, and
$N_{11}$ is the number of object pairs that are put into different clusters in 
$\mathcal{C}_t$ and also in different clusters in $\mathcal{C}$.
Note that values of all three measures range from 0 to 1, with a larger value indicating a better
clustering quality. 
}

\subsection{Algorithms for comparison}
\label{sec:algo-comp}
We first compare \conch\ with 11 other methods, which can be grouped into two categories.
Readers are referred to Section~\ref{sec:related} for more details.


\noindent
\textbf{(Network-embedding-based methods)}:
\textbf{node2vec} and \textbf{metapath2vec} (or mp2vec for short) are two network embedding methods that generate objects' embedding vectors.
These vectors are fed into a logistic regression classifier to predict objects' labels.
Since node2vec is designed for homogeneous networks,
we apply it to an HIN by ignoring the heterogeneity of the network.
For metapath2vec, 
we test all the given meta-paths and report the best result.

\noindent
\textbf{(Graph-neural-network-based methods)}:
\textbf{GCN} is a model that extends the convolution operation on graphs.
\textbf{GAT} further integrates the attention mechanism in the convolutional layer.
\textbf{MVGRL} is a self-supervised learning method.
All the methods are designed for homogeneous networks.
We apply them by converting an HIN to a homogeneous network using meta-paths and report the best result.
\textbf{HAN},
\textbf{HetGNN},
\textbf{MAGNN},
\textbf{HGT},
\textbf{HDGI} and 
\textbf{HGCN} are state-of-the-art heterogeneous neural network models.
Note that HAN, MAGNN and HDGI utilize meta-paths while others not.
In particular,
HDGI uses HAN as the base neural network model.
For MVGRL, HetGNN and HDGI, 
since they are unsupervised,
the learned object embeddings
are fed into a logistic regression classifier to predict objects' labels.

\subsection{Experimental setup}
\label{sec:setup}
We implement \conch\ using PyTorch. 
The model is initialized by Glorot initialization \cite{GlorotB10} and trained by Adam~\cite{KingmaB14}.
For \conch\ and all its variants,
we set
the learning rate to $0.001$,
the dropout ratio to $0.5$ and
the penalty weight on the $\ell_2$-norm regularizer to $0.0005$ on all three datasets.
We set $k$ in neighbor filtering to 5, 10, 10 and 
the number of layers $L$ to 2, 1, 1
in DBLP, Yelp and Freebase, respectively.
For the self-supervised regularization weight $\lambda$,
we fine tune it from $\{0.001,0.01,0.1,1\}$.
We use early stopping with a patience of 100, i.e.,
we stop training
if the validation accuracy does not decrease for 100 consecutive epochs.
To run \conch,
we first run metapath2vec with the default parameter settings in the original codes provided by the authors
to construct context features.
We set the initial context embedding dimensionality to 128 for DBLP and Yelp, and 32 for Freebase.
We perform neighbor filtering and context feature extraction as preprocessing steps.
For fair comparison,
we set the output embedding dimensionality for all the methods to be 128.
We run all the experiments on a server with 128G memory and a single NVIDIA 2080Ti GPU.
We also feed all the methods the same training/validation/test set splits.
For all the baseline methods, we use the original codes released by their authors.
For MVGRL, HetGNN and HDGI, since they are unsupervised,
we train the models by the train loss early stopping with 100 epochs.
For other graph neural network models,
we use the same early stopping criteria as \conch\ for fairness.
For these models,
we fine tune the model hyper-parameters by grid search. These include the number of model layers $\{1,2,3\}$
and the learning rate $\{0.01,0.001,0.0001\}$.
For HGCN, as suggested in the original paper,
we further perform the hyper-parameter tuning for 
the number of inception layers $\{1, 2, 3, 4\}$,
the convolutional kernel size $\{1, 2, 3, 4\}$
and the order of label propagation $\{0, 1, 2\}$.
For all the meta-path-based methods,
we employ the same meta-path sets.
For the random-walk-based embedding methods, 
we use the default parameters reported in their original papers.
For each method, we run experiments 10 times and report the average results.
We provide our code and data here:
\url{https://github.com/dingdanhao110/Conch}.

\subsection{Performance results}
\label{sec:results}

\begin{table*}[!htbp]
\centering
\caption{The classification results (\%) over the methods}
\label{tab:result}
\resizebox{\linewidth}{!}
{
\begin{tabular}{|c|c|c||c|c||c|c|c|c|c|c|c|c|c||c|}
\hline
Datasets                  & Metrics                   & Training & node2vec & mp2vec     & GCN & GAT & MVGRL & HAN & HetGNN & MAGNN & HGT & HDGI & HGCN  & \conch   \\ \hline
\multirow{8}{*}{DBLP}     & \multirow{4}{*}{Macro-F1} & 2\%     &    90.60   &    91.16       & 91.15  &  86.73  & 91.43 & 80.65  & 89.92 & 92.35 & 89.97  &   $67.71$ &  82.36 & \textbf{93.86}   \\
                          &                           & 5\%   &  92.29 &  92.46           & 91.86 &  90.59  & 91.66 &  87.74 & 91.99 & 93.01 &  92.00  & $85.68$ & 86.69   & \textbf{94.22}  \\
                          &                           & 10\%     &   92.08   &   92.48         & $92.20$ & $91.19$ & 91.99 & $90.98$  & 92.50 & 93.68 &  92.32 & $90.37$ &  88.36  & \textbf{93.77}    \\
                          &                           & 20\%   &  92.08   &   92.41          &$92.46$ & $91.23$ & 92.36 &  $92.51$  & 92.86 & 94.17 & 93.14  &  $91.60$ & 86.81  & \textbf{94.29}   \\ \cline{2-15} 
                          & \multirow{4}{*}{Micro-F1} & 2\%        &  91.54  &     91.88      & 91.94  & 88.27 &  92.01 & 82.30  & 90.92 & 92.99 & 90.87  & $77.33$ &  85.46  & \textbf{94.29}    \\
                          &                           & 5\%   & 92.82 &    93.00         & 92.56 & 91.20 & 92.26  & 88.38  & 92.59  & 93.58 &  92.62   & $88.10$ & 88.65   & \textbf{94.64} \\
                          &			     & 10\%  &  92.62   &   93.04          & $92.88$ & $91.77$ & 92.57  & $91.49$  & 93.01 & 94.13 & 92.90  & $91.35$ &  90.09  & \textbf{94.22}   \\
                          &                           & 20\%   &   92.65   &   92.95          &$93.10$ &  $91.78$ & 92.89 & $92.91$ & 93.36 & 94.56 &  93.65 &  $92.33$ &  88.67  & \textbf{94.70}  \\ \hline
\multirow{8}{*}{Yelp}     & \multirow{4}{*}{Macro-F1} & 2\%         &  75.54   &      84.95     & 84.80  & 70.34 &  53.77 & 58.12  & 84.86 & \multirow{4}{*}{-} & 78.36  &  $51.63$ &  59.07  & \textbf{88.60}   \\
                          &                           & 5\%   & 86.61  &       89.44     & 85.96 &  80.50 & 53.82 & 61.55  & 88.48 & & 88.64  & $56.83$ & 62.94  & \textbf{90.11} \\
    			 &  			     & 10\%  &   88.65   &   90.15          & 86.34& 80.78& 54.09 & 61.42 & 89.45 & & 90.00 &  $63.36$ &  73.86  & $\textbf{91.31}$  \\
                          &                           & 20\%  & 89.62   &   90.57          &86.39 & 81.95& 55.91 & 66.37 & 89.85 & & 90.32  & 72.59 & 76.61 & $\textbf{92.11}$  \\ \cline{2-15} 
                          & \multirow{4}{*}{Micro-F1} & 2\%     & 78.28  &   85.58         &  83.17  &  76.67  & 72.42 & 70.89 & 85.25 &  \multirow{4}{*}{-} & 80.78  & $65.53$ &  70.03  & \textbf{88.41}    \\
                          &                           & 5\%   & 85.43 &     88.31      & 84.09 & 81.30 & 73.15 & 72.71 & 87.51  & & 87.66  & $68.80$ &  72.40 & \textbf{89.69} \\
                          &			     & 10\%   &   87.50   &   89.10          & 84.53& 81.64 & 73.22 & 73.86  & 88.56 & & 89.10  & 72.64 & 78.58  &  $\textbf{90.78}$   \\
                          &                           & 20\%   &    88.45   &   89.56          & 84.58& 82.16 & 73.33 & 74.95 & 88.96 & & 89.48  & 77.51 & 79.73  & $\textbf{91.56}$   \\ \hline
\multirow{8}{*}{Freebase} & \multirow{4}{*}{Macro-F1}& 2\%     &  52.81   &    52.08       &  52.35  &  47.72 & 50.77 & 51.39 & 56.32 & 51.54 & 53.20 & $52.69$ &  47.73  & \textbf{56.46}   \\
                          &                           & 5\%      &  54.27   &    52.68       & 55.28 & 48.46 & 50.79 & 54.57 & 59.74 & 57.30 & 57.00  & 57.00&  51.65  & \textbf{61.07} \\
                          &			     & 10\%  &   55.38   &   54.26          & 59.08 & 50.00 & 51.14 & 57.96 & 62.15 & 59.50 & 60.39 & 59.93&  56.69  & \textbf{63.35}  \\
                          &                           & 20\%  &   57.35   &   57.50          & 60.90 & 51.09 & 51.15 & 60.11 & 62.92 & 62.23 & 61.76 & 61.39 &   60.35  & $\textbf{64.75}$  \\ \cline{2-15} 
                          & \multirow{4}{*}{Micro-F1}& 2\%        &  59.71  &    59.01       & 64.89  &  64.69 & 62.75 & 65.59 &  63.39 & 63.31 & 64.06 & 63.86 &  55.43 & \textbf{65.82}   \\
                          &                           & 5\%   &   59.67  &    57.83       & 66.00 &  65.53 & 64.44 & 65.83 & 65.40 & 65.65 & 66.09 & 65.41 &  58.79  & \textbf{67.55} \\ 
                          &			     & 10\%  &    60.19   &   58.91          & 67.65 & 65.86 & 65.52 & 66.79 & 67.41 & 66.93 & 67.69 & 66.82 &  63.03  & \textbf{69.21}  \\
                          &                           & 20\%  &   61.95   &   61.84          & 68.49 & 66.38& 65.92 & 67.32 & 68.05 & 68.41 & 68.85 & 67.55 &  66.75  & $\textbf{70.04}$   \\ \hline
\end{tabular}
}
\end{table*}

Table~\ref{tab:result} summarizes the performance results.
We evaluate the methods on three datasets under two evaluation metrics.
Moreover, to compare the methods when labeled objects are scarce, 
we vary the training set size from very small (2\% and 5\%) to moderate (10\% and 20\%).
There are thus in total 
3 $\times$ 2 $\times$ 4 = 24 ``contests'' in the comparison study.
Each row in the table corresponds to one contest.
For each contest,
we highlight the winner's score in bold.
From the table,
we make the following observations:

(1)
As the training set size decreases,
the performance of many baseline methods drops sharply.
For example,
on DBLP,
the Micro-F1 score of HDGI is $0.9233$ with 20\% labeled objects, which is close to the winner's score.
However, with only 2\% labeled objects,
the score of HDGI degrades to $0.7733$ while the best score (ConCH) is $0.9429$.


(2)
For the two network-embedding-based methods,
mp2vec performs better on DBLP and Yelp, while
node2vec takes the lead on Freebase. 
Although mp2vec is an embedding method for HINs that uses meta-paths,
it can take only one meta-path as input, which affects its performance.

(3)
\conch\ outperforms graph neural network models GCN and GAT.
This is because both methods are designed for homogeneous networks without considering objects' and links' heterogeneity. 
While MVGRL leverages self-supervision, it ignores heterogeneity and is outperformed by \conch.

(4)
Although HAN is designed for HINs and it uses meta-paths,
HAN omits mp-contexts,
which lowers its effectiveness.
For example, its Macro-F1 score on Yelp with $20\%$ labeled data is 0.6637, which is significantly lower than that of \conch.
This shows the importance of mp-contexts. 
Moreover,
since HDGI takes HAN as the base neural network model,
its performance is also adversely affected due to the ignorance of mp-contexts.
MAGNN considers mp-contexts and it performs well on DBLP.
However, as we will show in Sec.~\ref{sec:efficiency}, MAGNN is computationally expensive.
Further,
MAGNN needs to preprocess and maintain information for each meta-path instance, which requires a lot of storage.
For example, the meta-paths processing in the Yelp task generates a large number of path instances between objects,
which causes out-of-memory errors. As a result, in our experiments, MAGNN fails to run on Yelp.

(5)
HetGNN and HGT can achieve overall good performance over all three datasets.
However,
they are also easily affected when the number of labeled objects is limited.
For example,
with 2\% labeled objects,
the Macro-F1 scores of HetGNN and HGT on Yelp are $0.8486$ and $0.7836$, respectively, while the winner's is $0.8860$.
HGCN constructs relational features for an object by aggregating its neighbors' label information.
However, the constructed relational features and the original features of the object could be in different feature space,
which restricts the model's effectiveness.

(6) \conch\ achieves the best performance in all 24 cases.
\conch\ uses 
a simple neighbor filtering scheme to remove the less relevant neighbors of an object. 
Moreover, \conch, based on meta-paths, leverages mp-contexts to boost the classification performance.
We further observe that,
given scarce labeled objects,
\conch\  achieves superior performance compared with other methods.
For example,
with 2\% labeled objects,
the Micro-F1 score of \conch\ on Yelp is $0.8841$
while that of the first runner-up is only $0.8558$.
This is because
the self-supervision learned from unlabeled data
provides additional information that helps classify objects.

\subsection{Ablation study}
We conduct an ablation study on \conch\ to understand the characteristics of its main components. 
One variant updates objects' embeddings by directly aggregating information from its multi-hop neighbors specified by meta-paths without
considering contexts derived from the path instances. This helps us understand the importance of including
mp-contexts in object classification.
We call this variant \textbf{\conch\_nc} (\textbf{n}o \textbf{c}ontexts).
Another variant randomly selects $k$ meta-path-based neighbors from a given object as {\it relevant neighbors}.
This is in contrast to \conch, which selects relevant neighbors by picking the top-$k$ most similar path-based neighbors based on PathSim 
scores (see Section~\ref{subsec:fn}). 
We call this variant \textbf{\conch\_rd} (\textbf{r}an\textbf{d}om), which helps us evaluate the effectiveness of our neighbor-filtering strategy.
To show the importance of the self-supervised regularization,
we train the model with $\mathcal{L}_{sup}$ only and call this variant \textbf{\conch\_{su}} (\textbf{su}pervised).
Moreover,
while we formulate our problem as a multi-task learning objective,
we can also train the model by a \emph{pre-training \& fine-tuning} strategy.
Specifically,
we first train the model by using $\mathcal{L}_{ss}$ only and
then take the learned parameters as the initialization to the model that uses $\mathcal{L}_{sup}$ only.
This variant can be used to show the advantage of the multi-task learning framework
and we call it \textbf{\conch\_{ft}} (\textbf{f}ine-\textbf{t}uning).
Finally,
we consider a variant of \conch\ that assigns meta-paths equal weights without the attention mechanism.
We call this variant \textbf{\conch\_{ew}} (\textbf{e}qual \textbf{w}eights).

\begin{figure}[!htbp]
    \centering
    \subfigure[Macro-F1]{\includegraphics[width=0.22\textwidth]{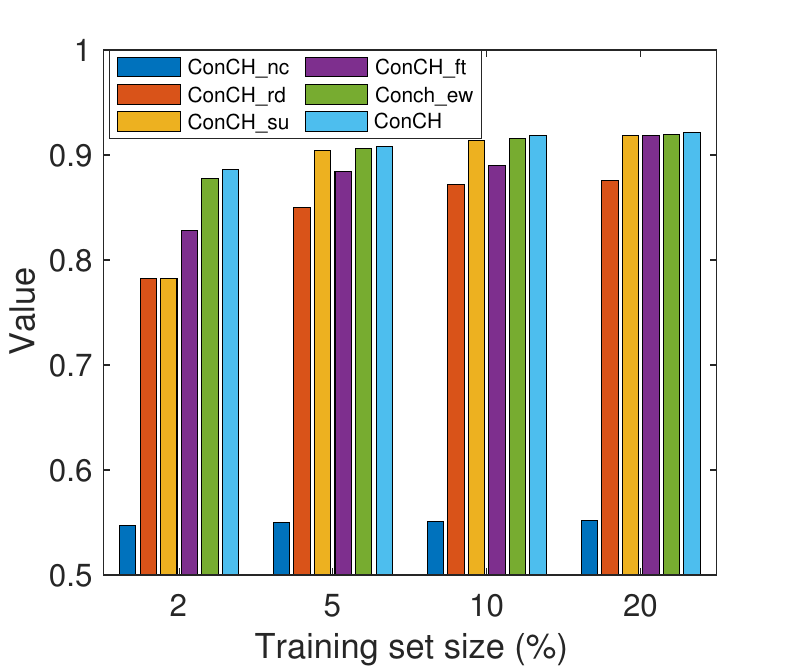}}    
    \subfigure[Micro-F1]{\includegraphics[width=0.22\textwidth]{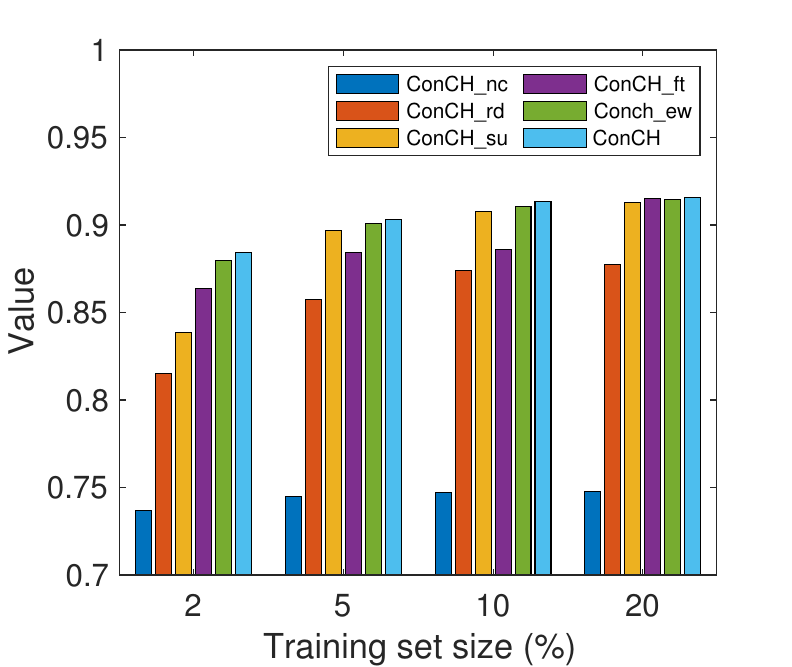}} 
     \caption{The ablation study results on Yelp}
     \label{figure:abl_yelp}
\end{figure}

\begin{figure}[!htbp]
    \centering
    \subfigure[Macro-F1]{\includegraphics[width=0.22\textwidth]{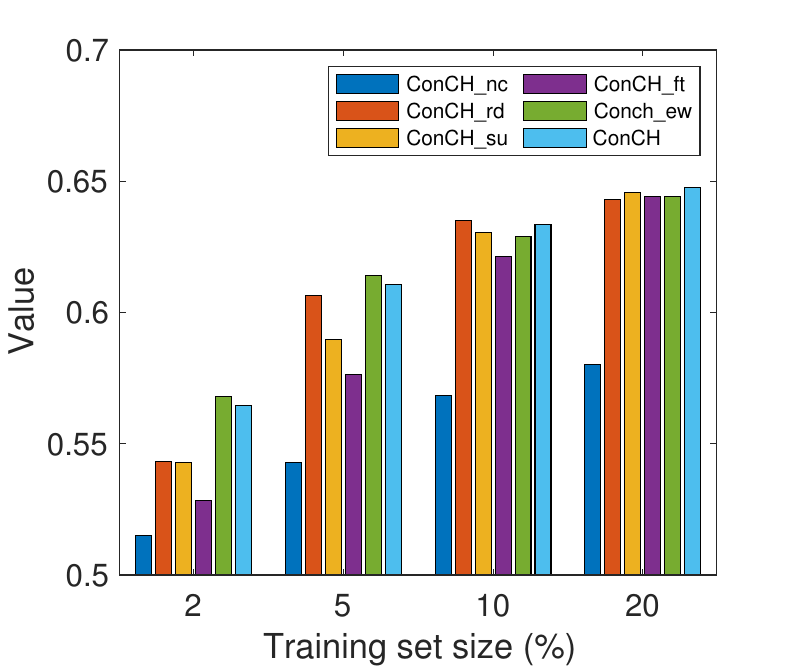}}    
    \subfigure[Micro-F1]{\includegraphics[width=0.22\textwidth]{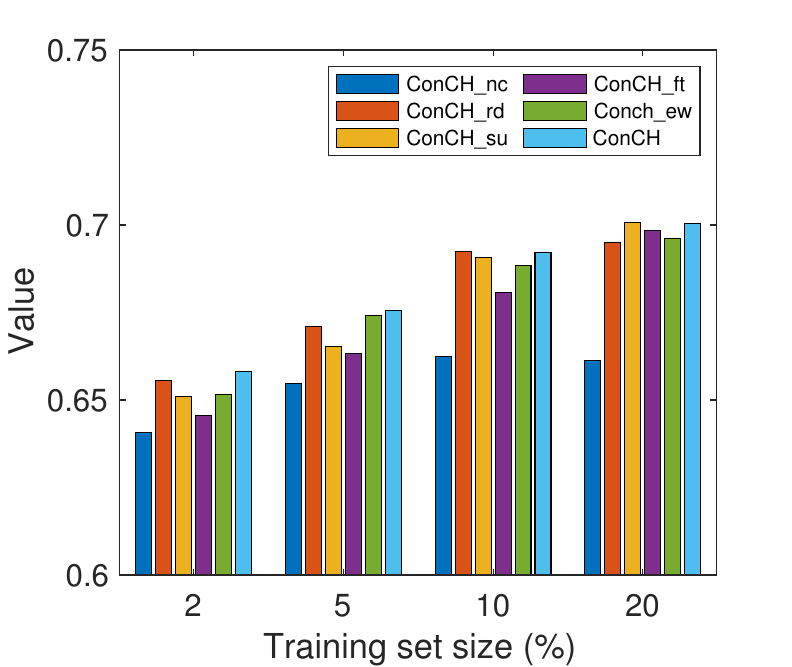}} 
     \caption{The ablation study results on Freebase}
     \label{figure:abl_freebase}
\end{figure}

Similar to Sec.~\ref{sec:results},
we compare \conch\ with these variants for
four training set sizes on three datasets w.r.t. Macro-F1 and Micro-F1 measures.
The results are given in Figs.~\ref{figure:abl_dblp} - \ref{figure:abl_freebase}.
From these figures,
we observe:

\begin{figure}[!htbp]
    \centering
    \subfigure[Macro-F1]{\includegraphics[width=0.22\textwidth]{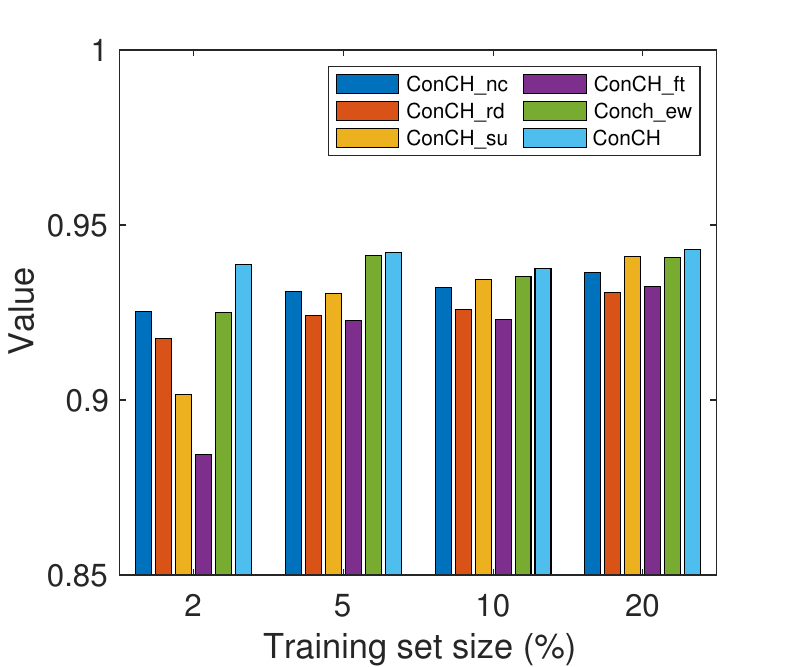}}    
    \subfigure[Micro-F1]{\includegraphics[width=0.22\textwidth]{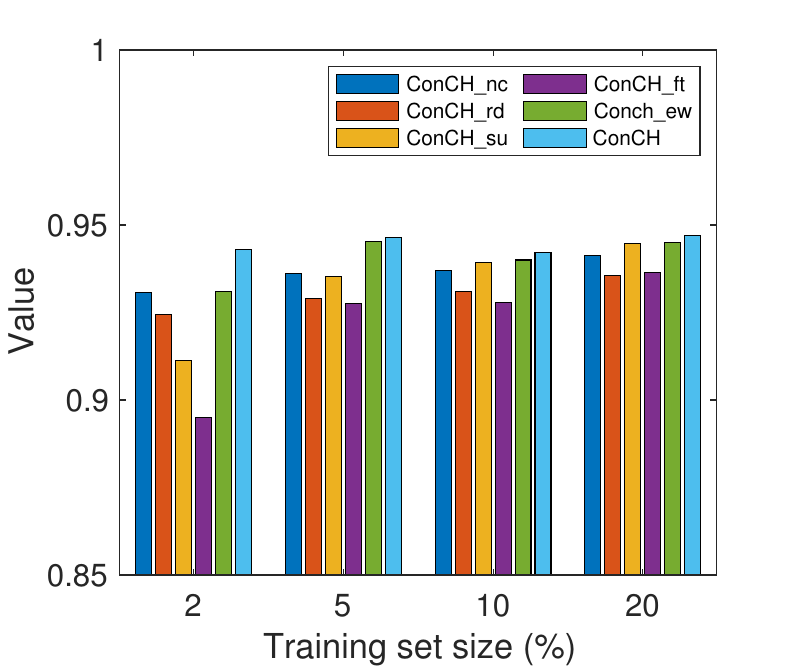}} 
     \caption{The ablation study results on DBLP}
     \label{figure:abl_dblp}
\end{figure}

(1)
\conch\ beats \conch\_nc in all 24 cases.
In particular, \conch\ significantly 
outperforms \conchnc\ on Yelp and Freebase.
This shows that mp-context information is particularly important for HINs
that are with heterogeneity in object types and link types.
When using meta-paths,
the inclusion of mp-contexts is essential for effective classification. 

(2)
\conch\ achieves better performance than \conchrd.
Since \conchrd\ randomly selects $k$ path-based neighbors for an object,
the performance gaps between \conch\ and \conchrd\
show that \conch's top-$k$ neighbor-filtering strategy is very effective in selecting more influential path-based neighbors
to improve classification accuracy.

(3)
Given 20\% labeled objects,
\conch\_su achieves comparable performance with \conch. 
However,
as the training set size decreases,
the performance gap between \conch\ and \conch\_su gets larger.
This further shows the importance of leveraging 
self-supervised learning to cover the shortage of labeled objects.

(4)
\conch\ clearly outperforms \conch\_ft on three datasets.
\conch\_ft, which switches the objective function from $\mathcal{L}_{ss}$ in pre-training to $\mathcal{L}_{sup}$ in fine-tuning,
solves two optimization problems. 
However,
\conch\ unifies the two loss functions into one objective
that is much easier to be optimized and can better leverage both labeled and unlabeled objects.
Our results are similar to those reported in~\cite{you2020does}.

(5)
\conch\ performs better than \conch\_ew in 22 cases.
This further shows the importance of the attention mechanism in learning meta-path weights.

\subsection{Attention weight learning}
\begin{figure}[!htbp]
    \centering
    \subfigure[DBLP]{\includegraphics[width=0.15\textwidth]{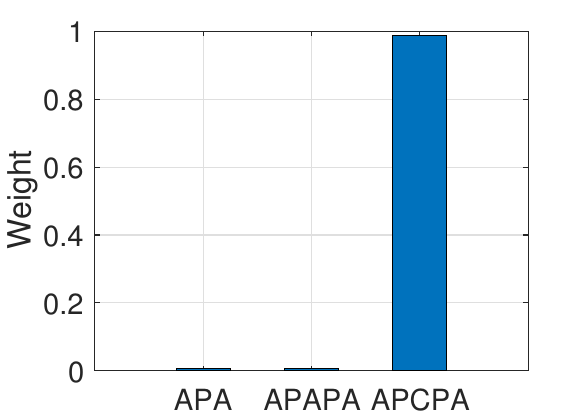}}    
    \subfigure[Yelp]{\includegraphics[width=0.15\textwidth]{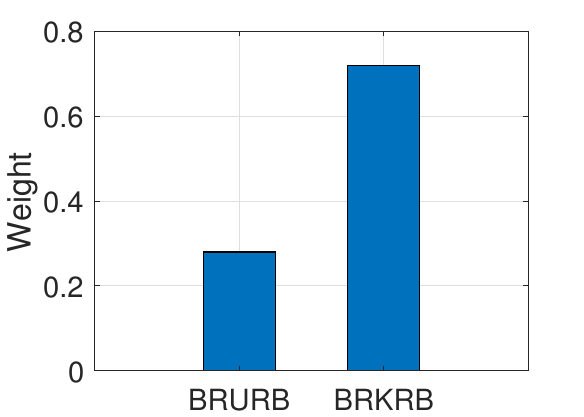}} 
    \subfigure[Freebase]{\includegraphics[width=0.15\textwidth]{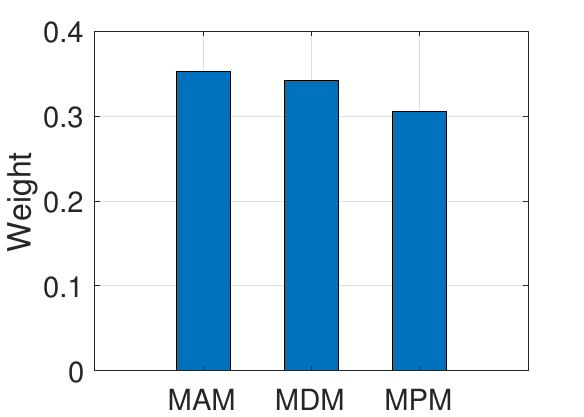}} 
     \caption{Learned weights of meta-paths}
     \label{figure:mp}
\end{figure}

\conch\ learns the importance of meta-paths by the attention mechanism.
To show the effectiveness of attention weight learning,
we compute the average learned weights of meta-paths for all three datasets.
We show the results (Fig.~\ref{figure:mp}) with $20\%$ labeled objects as training sets for illustration.


Fig.~\ref{figure:mp}(a) shows the learned meta-paths' weights for the DBLP dataset.
Recall that the classification task is to classify authors by their research areas.
From the figure, 
we see that
the weight of the meta-path APCPA (authors that publish papers in the same venue) is almost $1$,
while
that of meta-paths APA (co-authorship) and APAPA (authors that share a common co-author) are very close to $0$.
An almost-0 weight for APA may seem counter-intuitive considering that 
 co-authorship is a strong signal that two authors work in the same area.
The reason why APA is given such a low weight
is that it is a sparse relation.
An author typically co-authors with only a handful of others in the community. 
Further,
an author is related to a large number of other authors by 
APCPA (conference co-attendant relation).
Moreover, authors related by APA are also related by APCPA, hence, the former is subsumed by the latter.
In this example, we see that 
\conch\ correctly selects the relevant meta-path APCPA over APA and APAPA. 

Fig.~\ref{figure:mp}(b) shows the learned meta-paths' weights for the Yelp dataset.
Recall that the task is to classify restaurants by their food categories.
We see that the meta-path 
BRKRB (restaurants whose reviews contain the same food keyword)
is given a much larger weight than BRURB (restaurants that are visited by the same customer).
This is reasonable because food keywords in reviews directly indicate  food category.
On the other hand,
a customer could visit restaurants that serve different categories of food.

The Freebase task is to classify movies by genre.
Fig.~\ref{figure:mp}(c) shows
that all three meta-paths are useful.
The weights of
meta-paths MAM (movies share the same actor)
and MDM (movies filmed by the same director)
are a bit larger than that of MPM (movies produced by the same producer).
From our discussion,
we see that \conch's meta-path weight learning strategy is highly effective.


\begin{figure*}[!htbp]
    \centering
    \subfigure[DBLP]{\includegraphics[width=0.22\textwidth]{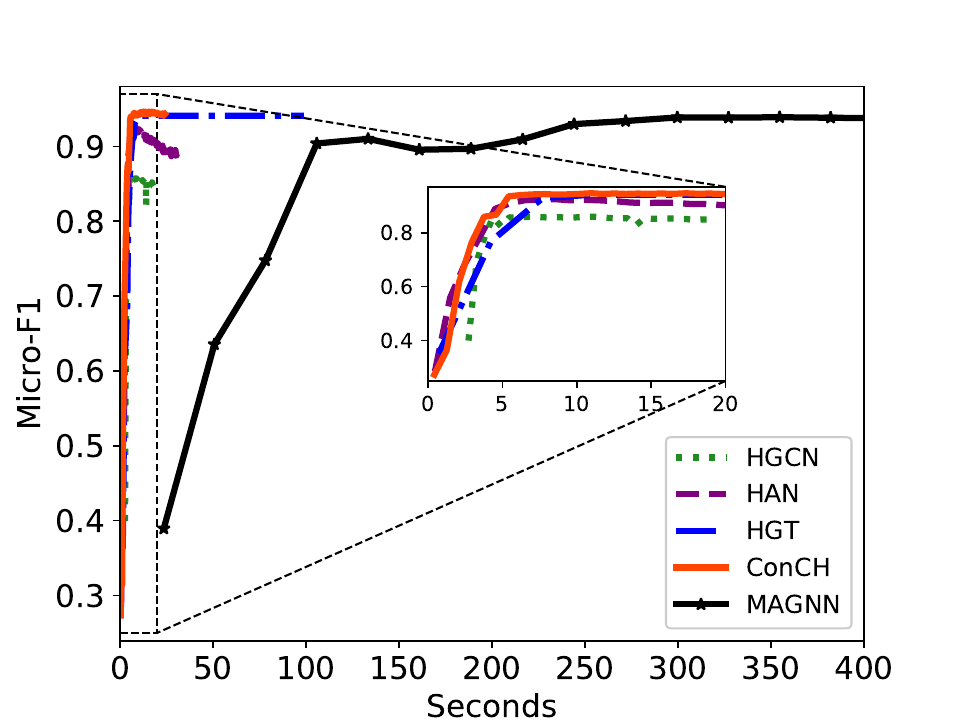}}    
    \subfigure[Yelp]{\includegraphics[width=0.22\textwidth]{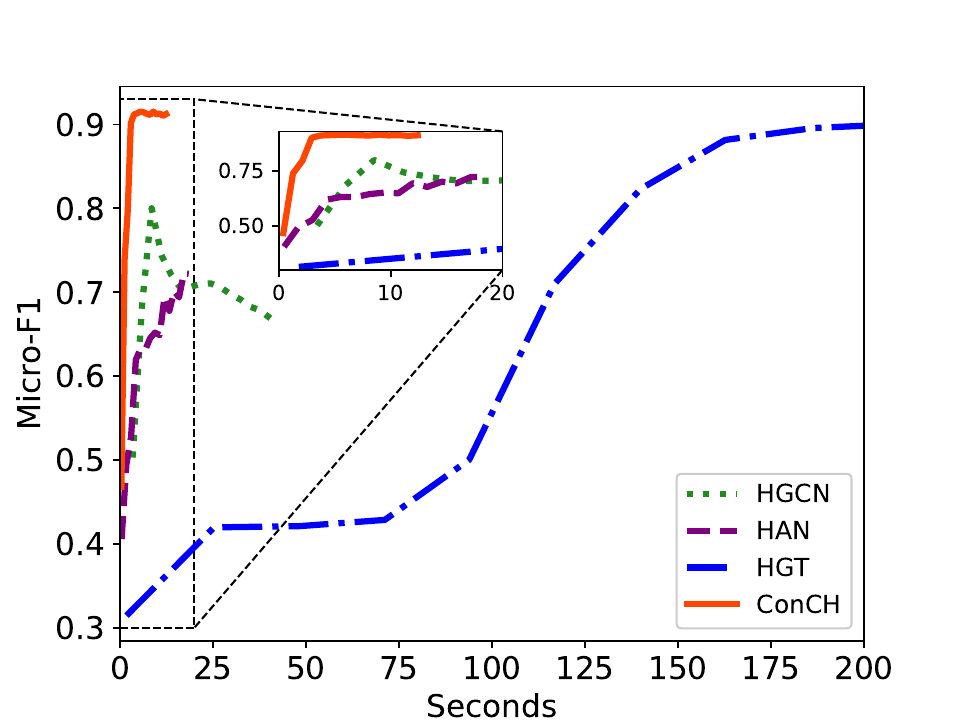}} 
    \subfigure[Freebase]{\includegraphics[width=0.22\textwidth]{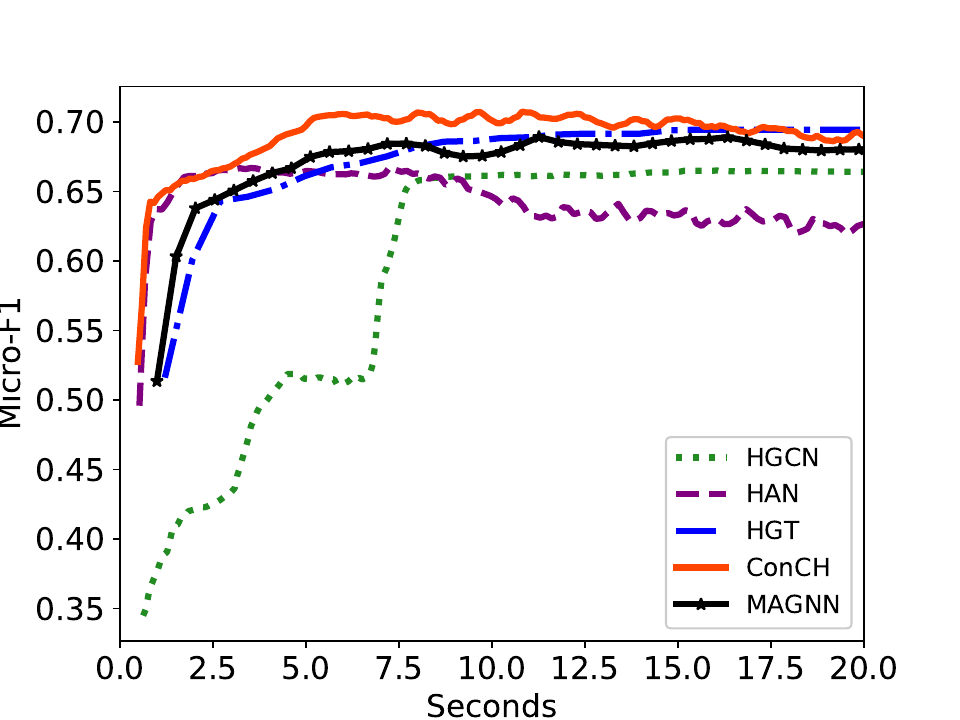}} 
     \subfigure[One training epoch runtime vs. $k$]{\includegraphics[width=0.22\textwidth]{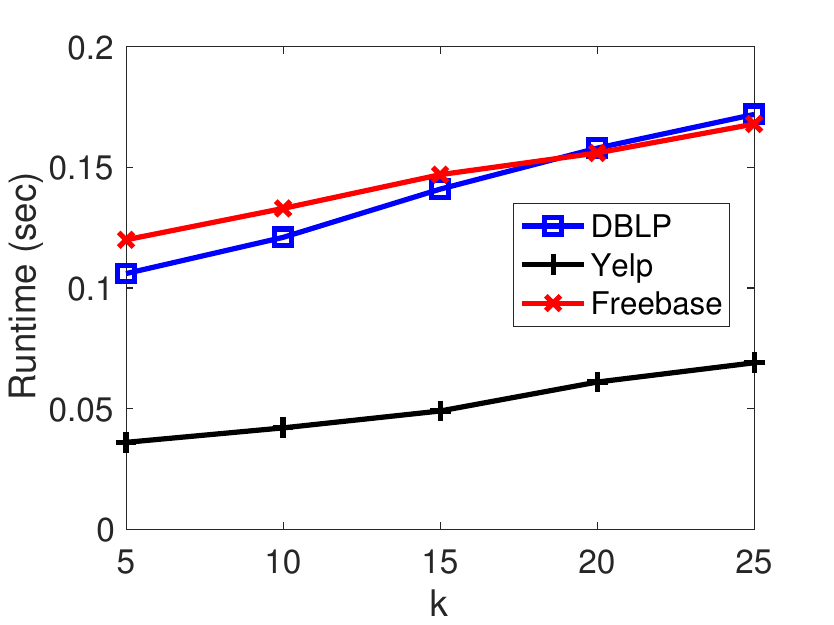}} 
     \caption{The results of efficiency study. Specifically, (a)-(c) compare the training time vs. the Micro-F1 score on three datasets; (d)
     shows the runtime of one training epoch in \conch\ vs. $k$ on three datasets}
     \label{figure:runtime}
\end{figure*}

\subsection{Efficiency study}
\label{sec:efficiency}
In this section we study \conch's efficiency.
For fairness,
we compare the training time for \conch, HAN, MAGNN, HGT and HGCN,
as they are semi-supervised classification methods in HINs.
For all these methods, we use the same training/validation set and run the experiments for 300 epochs.
We take 20\% labeled objects as training set for illustration.
We repeat all the experiments three times and 
show the average training time of these methods w.r.t. the Micro-F1 score on the validation set in Fig.~\ref{figure:runtime}.
Note that we cannot run MAGNN on Yelp.
To recap the major differences of these methods:
\conch, HAN and MAGNN are based on meta-paths.
However,
\conch\ selects the $k$ most informative path-based neighbors for an object based on a filtering scheme.
HAN uses all meta-path-based neighbors (i.e., no filtering) and
it learns the neighbors' relative importance by the vanilla attention mechanism. 
HAN does not use mp-contexts either.
Different from \conch,  which uses mp-contexts to construct high-level context objects with features,
MAGNN utilizes mp-contexts 
in a fine-grained manner by independently considering all the meta-path instances.
It learns the importance of meta-path instances and further aggregates information from these path instances to generate object embeddings. 
HGT employs a Transformer based aggregator to aggregate information from multi-type neighbors for an object,
while 
HGCN leverages multiple kernels that use different convolutional filters.

From Figs.~\ref{figure:runtime}(a)-(c), we can see that 
\conch\ consistently converges fast to the best results over all three datasets.
MAGNN and HGT can also reach high performance scores,
however,
they need much longer time to converge.
In particular, for MAGNN, 
when meta-paths are of longer lengths,
the number of path instances could be numerous,
which adversely affects model efficiency.
For example,
\conch\ is about $50 \times$ faster than MAGNN on DBLP;
\conch\ achieves almost $ 40 \times$ speedup than HGT on Yelp.
Although HAN and HGCN
are faster than MAGNN and HGT,
their performances are poor.
These results show that our approach \conch\ is 
very efficient and also highly effective.
Finally, the runtime of \conch\ depends on $k$,
which is the number of selected relevant neighbors for an object.
Fig.~\ref{figure:runtime}(d) shows how \conch's runtime of one training epoch varies with $k$.
From the figure, we see that the runtime increases fairly linearly with $k$.
Thus, \conch\ is scalable w.r.t. $k$. 

To further study \conch's efficiency,
we conduct experiments on 
a large dblp-4area subgraph extracted from the AMiner citation network~\footnote{https://originalstatic.aminer.cn/misc/dblp.v12.7z}.
We first extract papers whose fields of study in the original dataset contain at least one of four areas:
database (DB), data mining (DM), machine learning (ML), and information retrieval (IR).
For simplicity, we only consider conference papers.
We extract authors and conferences related to these papers.
The resulting dataset contains 416,554 papers (P), 537,435 authors (A) and 2,649 conferences (C).
Links include A-P (an author publishes a paper) and P-C (a paper is published at a conference).
Our task is to classify papers into one of the four research areas.
For each paper, following~\cite{yang2020relation},
we compute a 300-dimension attribute vector by averaging the word embeddings of keywords it contains.
We consider the meta-path set \{PAP, PCP\}.
The ground truth labels of papers are derived from the original dataset,
where we label each paper by the research area with the largest field of study weight.

\begin{table*}[t]
\centering
\caption{The classification results (\%) over all the methods on AMiner}
\label{table:aminer_result}
\resizebox{\linewidth}{!}
{
\begin{tabular}{|c|c|c||c|c||c|c|c|c|c|c|c|c|c||c|}
\hline
Datasets                  & Metrics                   & Training & node2vec & mp2vec     & GCN & GAT & MVGRL & HAN & HetGNN & MAGNN & HGT & HDGI & HGCN  & \conch   \\ \hline
\multirow{8}{*}{AMiner}     & \multirow{4}{*}{Macro-F1} & 2\%     &  57.06   &   33.89   &  59.91 & 68.25 & \multirow{4}{*}{-}  & 71.47  & 66.49 &  \multirow{4}{*}{-}  & 68.44  &  72.38   &  52.23 &  $\bm{73.10}$  \\
                          &                           & 5\%   & 57.19  &   33.91    &    60.24   & 70.66  &  &  73.87    & 66.84  &   &  69.01   & $73.24 $ &   49.39  &   $\bm{76.05}$ \\
                          &                           & 10\%     &    57.25   &  34.00  & 60.41  & $ 71.60$ & & $ 74.70 $ & $ 66.96$  &     &   69.37 & $ 73.62$ &   51.27  &   $\bm{77.11}$    \\
                          &                           & 20\%   &   57.26   & 38.33     &   60.50  &$ 72.00$ & & $ 75.44$ & $67.07$   &   &  69.72  &  $73.87$ & 53.94   & $\bm{78.33}$  \\ \cline{2-15}                
                          & \multirow{4}{*}{Micro-F1} & 2\%    &  65.77  & 52.68    &      66.83      &  72.79  &   \multirow{4}{*}{-}  &  75.79  & 72.07  &  \multirow{4}{*}{-}  & 72.50   & $76.88$ &   63.96  &  $\bm{77.54}$  \\
                          &                           & 5\%   & 65.89  &    52.72   &   67.11    &  74.91 &  & 77.92  &   72.35    &   &   73.18   & $77.67$ &  64.58   &  $\bm{79.95}$ \\
                          &			     & 10\%  &  65.95    &    52.71    &   67.29   & $ 75.84 $ & & $78.66$ & $72.41 $    &   & 73.56   & 77.92&   65.28  &  $\bm{80.85} $ \\
                          &                           & 20\%   &   65.95    &   54.57   &   67.37     &$ 76.22$ &  & $79.16$ & $72.46$ &   &  73.92  &  $78.09$ &  65.16   &  $\bm{81.84}$  \\ \hline                   
\end{tabular}
}
\end{table*}

\begin{figure}
    \centering
    \includegraphics[width=0.3\textwidth]{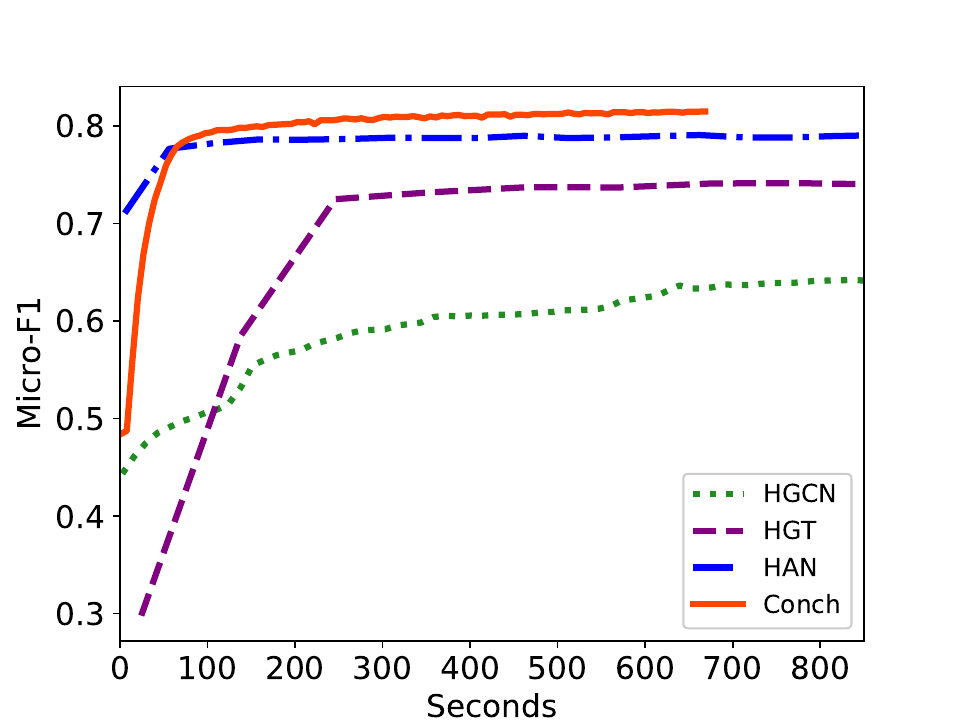} 
     \caption{Runtime comparison on AMiner}
     \label{figure:aminer}
\end{figure}

Figure~\ref{figure:aminer} shows the efficiency study on the AMiner dataset.
All these methods are semi-supervised classification methods in HINs.
Compared with others,
\conch\ converges fast to the best performance.
Table~\ref{table:aminer_result} further summarizes the classification performance over all the methods on AMiner.
\conch\ clearly outperforms all the baseline methods,
which again shows its effectiveness.
Due to the large dataset size,
both MVGRL and MAGNN cause out-of-memory errors.
Note that MVGRL
requires both adjacency matrix and diffusion matrix as input
to
provide both local and global views of a graph structure, respectively.
However,
the diffusion matrix computed by all the approaches suggested in the original paper is too dense,
which causes out-of-memory errors.
This further shows that 
\conch\ is efficient and also very effective.

\begin{figure*}[!htbp]
    \centering
    \subfigure[]{\includegraphics[width=0.22\textwidth]{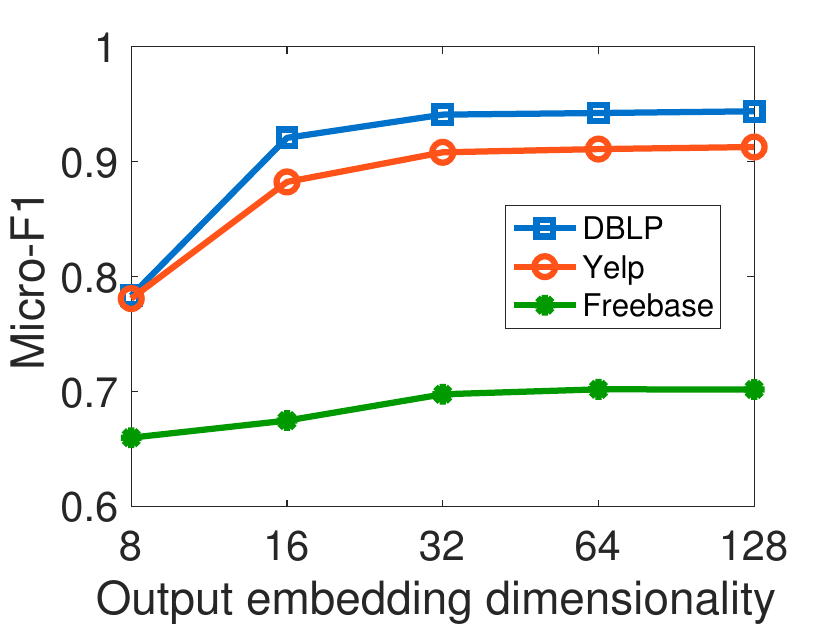}}    
    \subfigure[]{\includegraphics[width=0.22\textwidth]{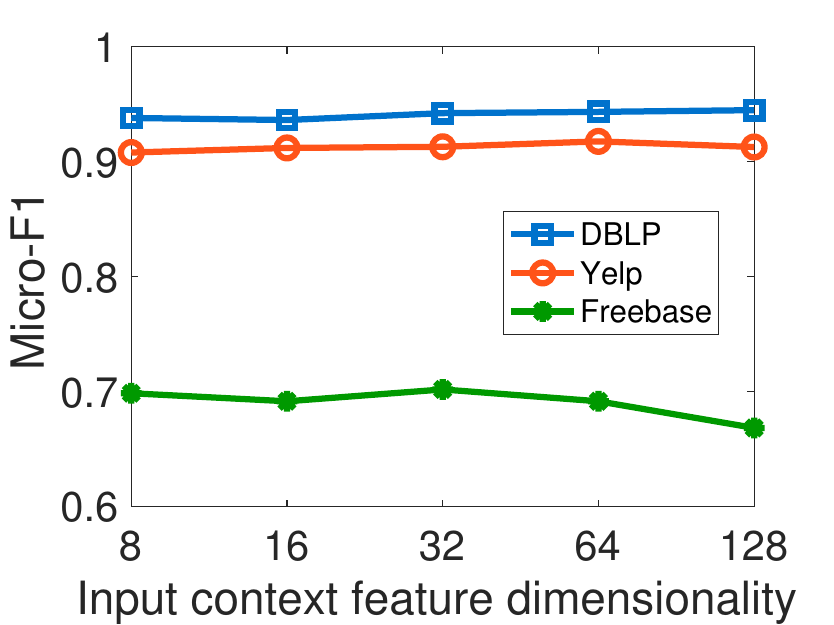}} 
    \subfigure[]{\includegraphics[width=0.22\textwidth]{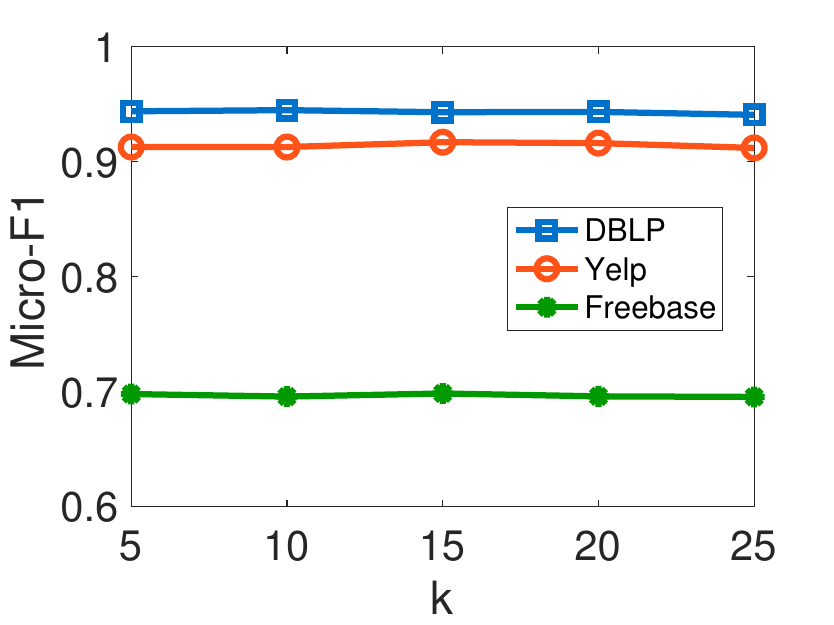}} 
    \subfigure[]{\includegraphics[width=0.22\textwidth]{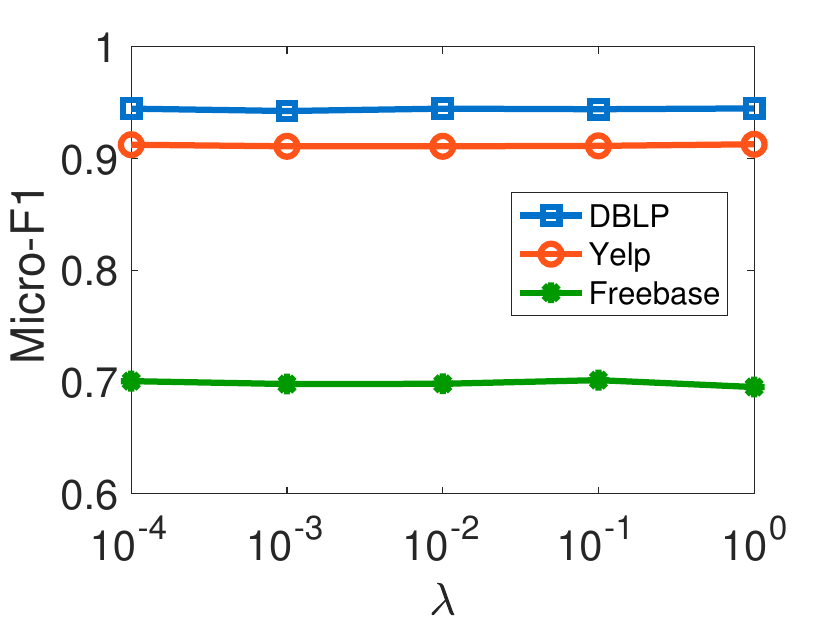}} 
     \caption{Parameter analysis}
     \label{figure:para}
\end{figure*}

\subsection{Parameter analysis}
We end this section with a sensitivity analysis on the hyper-parameters of \conch.
In particular,
we study four key hyper-parameters:
the output embedding dimensionality, 
the input context embedding dimensionality, the number of selected relevant neighbors $k$
and the self-supervised regularization weight $\lambda$.
In our experiments,
we vary one parameter each time with others fixed.
Fig.~\ref{figure:para} illustrates the results with 20\% labeled objects w.r.t. the Micro-F1 scores.
(Results on Macro-F1 scores exhibit similar trends, and thus are omitted for space reasons.)
From the figure, 
we see that

(1) As the output embedding dimensionality increases,
\conch\ achieves better performance. 
This is because when the dimensionality is small,
node embeddings cannot capture enough information for classifying objects.

(2)
There is a performance drop in Freebase when the input context feature dimensionality is 128.
This shows that 
initial context embedding vectors in a large dimensionality
could contain noise that adversely affects the classification accuracy.

(3)
For the other two hyper-parameters,
\conch\ gives very stable performances over a wide range of  parameter values.
In particular, 
\conch\ performs well even with small $k$.
This further shows that 
\conch\ is an effective and efficient method for solving the classification problem on HINs.

\section{Conclusions}
\label{sec:conclusion}
In this paper
we studied classification in HINs and 
proposed \conch,
which is a graph neural network model based on meta-paths.
\conch\ 
formulates the classification problem as a multi-task learning problem that combines semi-supervised learning
with self-supervised learning to enhance the model performance when the training set size is small.
Further,
it leverages meta-path-based contexts, which capture specific details of meta-path instances,  to improve classification accuracy.
It filters less relevant neighbors of an object by selecting the top-$k$ neighbors using PathSim scores. 
This approach helps reduce the number of neighbors whose information is aggregated to derive an object's embedding.
We conducted extensive experiments and compared \conch\ with 15 other methods.
Our analysis shows that \conch\ can achieve superior classification performance with high efficiency.


\bibliographystyle{IEEEtran}

\bibliography{classification}

\end{document}